\documentclass{article}
\usepackage[bb=px]{mathalpha}
\usepackage{natbib}
\usepackage{arxiv}

\usepackage[utf8]{inputenc} 
\usepackage[T1]{fontenc}    
\usepackage{url}            
\usepackage{booktabs}       
\usepackage{amsfonts}       
\usepackage{nicefrac}       
\usepackage{microtype}      
\usepackage{lipsum}
\usepackage[pdftex]{graphicx}
\graphicspath{ {./images/} }
\usepackage[english]{babel}
\usepackage{booktabs}
\usepackage{bbm}
\usepackage{amsmath}
\usepackage{graphicx}

\usepackage[acronym]{glossaries}
\makeglossaries

\usepackage{pdflscape}
\usepackage{longtable}

\usepackage{subcaption}
\usepackage{graphicx}
\usepackage{tabularx}
\usepackage{multirow} 
\usepackage{natbib}
\usepackage{amsmath,amssymb}

\usepackage{svg}

\usepackage{pdflscape}

\newacronym{mae}{MAE}{mean absolute error}

\newacronym{mape}{MAPE}{mean absolute percentage error}

\newacronym{rmse}{RMSE}{root mean squared error}

\newacronym{smape}{SMAPE}{symmetric mean absolute percentage error}

\newacronym{mse}{MSE}{mean squared error}

\newacronym{poi}{POI}{point of interest}

\newacronym{ignnk}{IGNNK}{Inductive Graph Neural Networks for Spatiotemporal Kriging}

\newacronym{gnn}{GNN}{Graph Neural Network}

\newacronym{zinb}{ZINB}{Zero Inflated Negative Binomial}

\newacronym{nb}{NB}{Negative Binomial}

\newacronym{gnll}{GNLL}{Gaussian Negative Log-Likelihood}

\newacronym{kl}{KL-divergence}{Kullback-Leibler divergence}

\newacronym{gnnui}{GNNUI}{Graph Neural Network for Urban Interpolation}

\newacronym{stgnn}{STGNN}{Spatio Temporal Graph Neural Network}

\newacronym{nll}{NLL}{Negative Log Likelihood}

\newacronym{ig}{IG}{integrated gradients}

\usepackage{nameref}
\usepackage{etoolbox}

\usepackage[colorlinks=true, linkcolor=blue, citecolor=blue, urlcolor=blue]{hyperref}

\usepackage[verbose=true,letterpaper]{geometry}
\AtBeginDocument{
  \newgeometry{
    textheight=9in,
    textwidth=6.5in,
    top=1in,
    headheight=14pt,
    headsep=25pt,
    footskip=30pt,
    left=1.5in,       
    right=1.5in,      
  }
}

\title{Spatio-Temporal Graph Neural Network for Urban Spaces: Interpolating Citywide Traffic Volume}

\title{Spatio-Temporal Graph Neural Network for Urban Spaces: Interpolating Citywide Traffic Volume}

\author{
  Silke K. Kaiser\textsuperscript{a,b,*},
  Filipe Rodrigues\textsuperscript{c},
  Carlos Lima Azevedo\textsuperscript{c},
  Lynn H. Kaack\textsuperscript{a,b} \\
  \textsuperscript{a}Data Science Lab, Hertie School, Berlin, Germany \\
  \textsuperscript{b}Centre for Sustainability, Hertie School, Berlin, Germany \\
  \textsuperscript{c}Department of Technology, Management and Economics, \\
  \quad Intelligent Transportation Systems, Technical University of Denmark \\
  \textsuperscript{*}Corresponding author: \texttt{s.kaiser@phd.hertie-school.de}
}

\begin{document}
\date{}
\maketitle
\begin{abstract}

Graph Neural Networks have shown strong performance in traffic volume forecasting, particularly on highways and major arterial networks. 
Applying them to urban settings, however, presents unique challenges: urban networks exhibit greater structural diversity, traffic volumes are highly overdispersed with many zeros, the best way to account for spatial dependencies remains unclear, and sensor coverage is often very sparse.
We introduce the Graph Neural Network for Urban Interpolation (GNNUI), a novel urban traffic volume estimation approach. GNNUI employs a masking algorithm to learn interpolation, integrates node features to capture functional roles, and uses a loss function tailored to zero-inflated traffic distributions. In addition to the model, we introduce two new open, large-scale urban traffic volume benchmarks, covering different transportation modes: Strava cycling data from Berlin and New York City taxi data. 
GNNUI outperforms recent, some graph-based, interpolation methods across metrics (MAE, RMSE, true-zero rate, Kullback-Leibler divergence) and remains robust from 90\% to 1\% sensor coverage. 
For example, on the Strava dataset, the MAE increases only from 7.1 to 10.5, and on the Taxi dataset, from 23.0 to 40.4. These results demonstrate that GNNUI maintains strong performance despite extreme data scarcity, a common condition in real-world urban settings.
We also examine how graph connectivity choices influence model accuracy.
\end{abstract}


\section{Introduction}
Reliable data on citywide traffic volumes is critical for transportation and urban planning decisions. Urban traffic data across modes can provide insights for targeted infrastructure improvements, traffic management, or improving the efficiency of public transportation \citep{leduc_road_2008, zheng_urban_2014}. Traffic volume data is collected by sparsely distributed physical traffic sensors that count the number of vehicles passing by. Extending this monitoring to all street segments (a street section between two intersections) is largely infeasible due to the high costs associated with deploying and maintaining such devices. We explore how traffic volumes in locations without sensors can be interpolated employing \acrlong{gnn}s (\acrshort{gnn}s), using datasets with full spatial and temporal coverage, taxi trip and Strava cycling counts, as proxies for overall traffic activity, given the limited availability of comprehensive ground-truth data.

Urban traffic volume is spatio-temporal, following a network-based structure. \acrshort{gnn}s have gained significant attention for traffic volume prediction due to their ability to model non-Euclidean spatial dependencies \citep{jiang_graph_2022}. When combined with temporal learning mechanisms, these models hold great promise for various urban applications \citep{jin_spatio-temporal_2024}.

\begin{figure}[ht!]
    \centering
        \begin{subfigure}{\textwidth}
        \centering
        \includegraphics[width=0.92\textwidth]{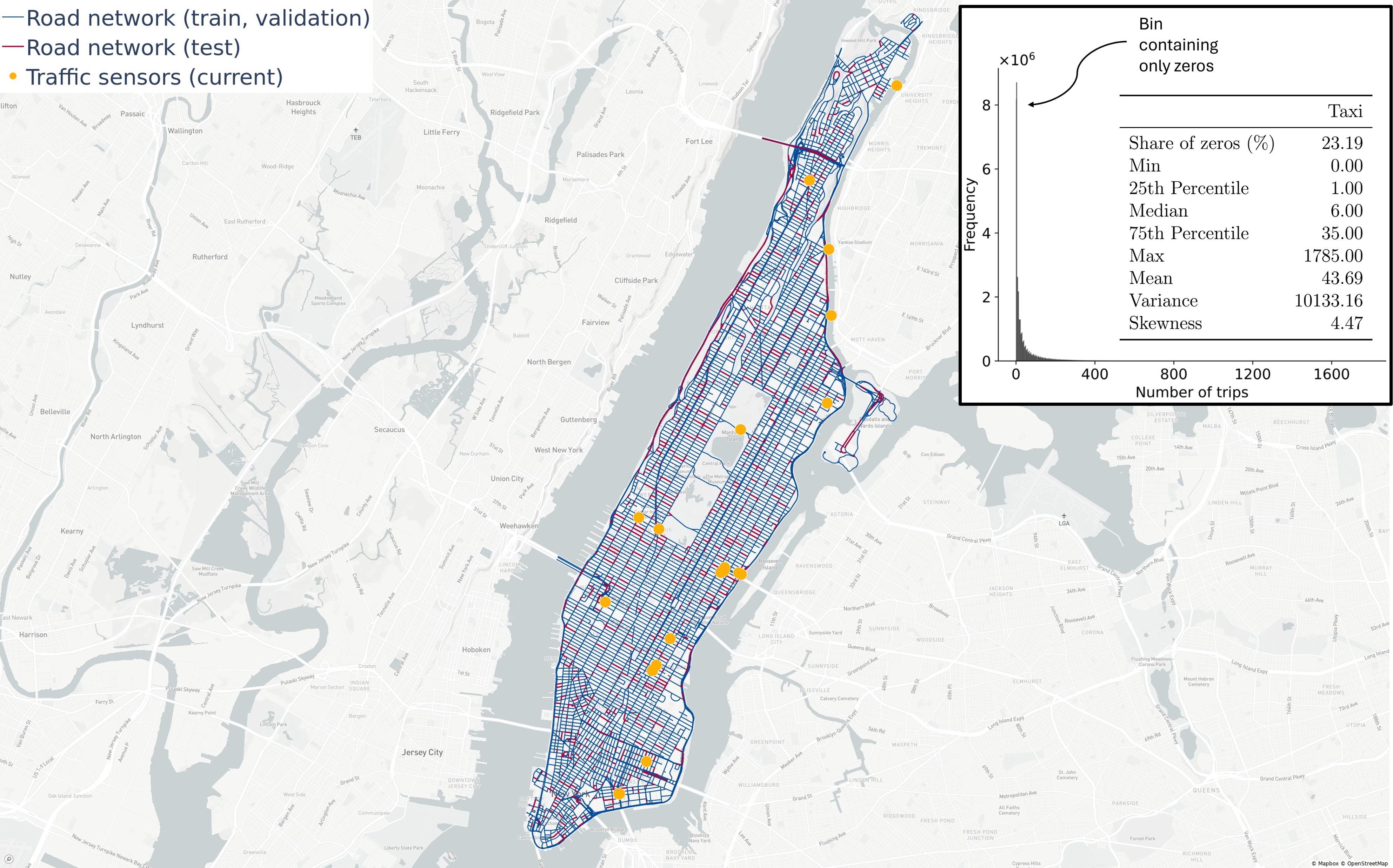}
        \caption{Street network of Manhattan, New York City, comprising 8,156 street segments. The 21 existing traffic sensors are shown as dots. The inset in the upper right displays the distribution of taxi trip volumes per street segment.}
        \label{fig:network_comparison_ny}
    \end{subfigure}
    
     \vspace{0.5cm}
     
    \begin{subfigure}{\textwidth}
        \centering
        \includegraphics[width=0.92\textwidth]{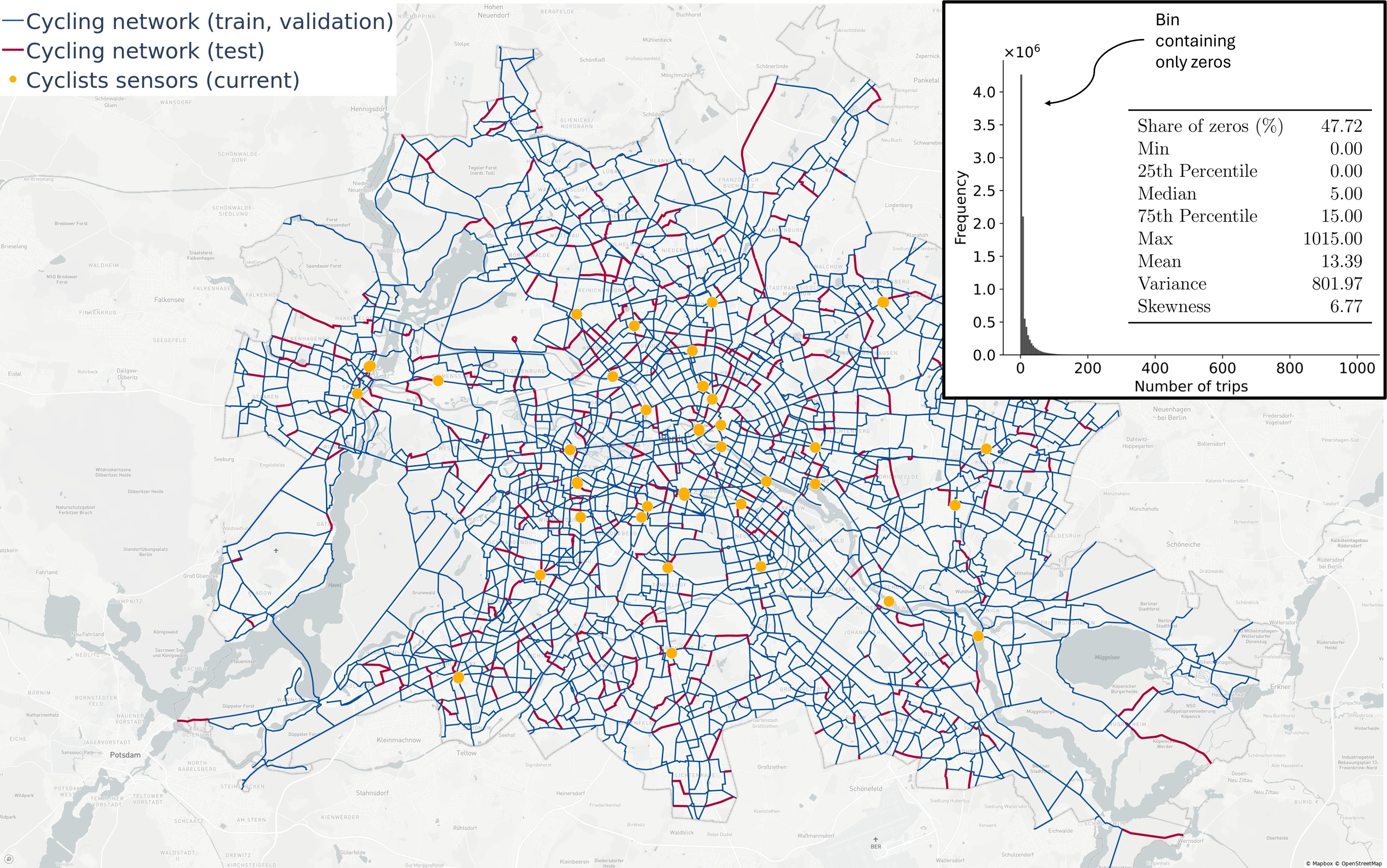}
        \caption{Cycling network of Berlin, comprising 4,958 street segments. The 41 existing bicycle sensors are shown as dots. The inset in the upper right displays the distribution of Strava cycling volumes per street segment.}
        \label{fig:network_comparison_berlin}
    \end{subfigure}
    \caption{Urban street networks (left) and corresponding distributions of street-level traffic volume (right). Street segments used for training and validation are shown in one color, and test segments in another, illustrating the spatial split applied in our experiments. Dots indicate the locations of the currently employed traffic and cycling sensors, which are displayed for reference only. Our analyses do not rely on these sensors; instead, we use trajectory-based data from taxi trips (New York) and Strava cycling records (Berlin). The sensors are shown solely to illustrate the sparsity of conventional urban monitoring infrastructure.}
    \label{fig:network_comparison_combined}
\end{figure}

Previous work has shown that interpolation with \acrshort{gnn}s performs well for highways and similar networks \citep{yao_spatiotemporal_2023, zheng_increase_2023, wu_inductive_2021, wu_spatial_2021, appleby_kriging_2020}. However, these networks differ profoundly from urban street networks, and correspondingly, the existing models overlook several critical aspects of urban contexts. 
Urban street networks exhibit greater infrastructural diversity than highways, making it essential to incorporate additional features that capture these differences. For example, the number of shops or the speed limit can influence traffic volumes. 
Moreover, the urban traffic data can be highly overdispersed, as in the case of specific transportation services and modes (e.g., cycling), with a significant proportion of zero values (Figure \ref{fig:network_comparison_combined}). Most existing models rely on the assumption of normally distributed data and adopt corresponding loss functions, which may not work for urban data. 
Furthermore, the higher density and structural complexity of urban networks complicate spatial dependency modeling. In poly-centric urban forms, for example, similar traffic patterns can emerge even between distant areas. Since \acrshort{gnn}s are highly sensitive to graph construction, a systematic comparison of graph configurations is needed.
Lastly, urban networks tend to exhibit far lower sensor density relative to the number of street segments when compared to highways. For instance, Manhattan’s street network comprises 8,156 segments, yet only 21 sensors recorded motorized traffic in 2023 \citep{nyc_open_data_automated_2024}. Berlin's cycling network spans 4,958 segments, but cyclist volume was collected by just 22 permanent and 19 temporary sensors \citep{senate_department_for_the_environment_mobility_consumer_and_climate_protection_berlin_radverkehrszahlstellen_2024} (see Figure~\ref{fig:network_comparison_combined}). This is in stark contrast to highway networks, such as California’s highway network, commonly used through the PeMS benchmark dataset, with over 40,000 sensors \citep{state_of_california_pems_nodate}. 

To address these challenges, we propose \acrlong{gnnui} (\acrshort{gnnui}), a model that interpolates traffic volumes while accounting for urban-specific characteristics. It captures spatial dependencies through graph structure. To model temporal dependencies and enable effective interpolation, we adopt a masking algorithm and time window approach from the IGNNK model by \cite{wu_inductive_2021}. To better capture the heterogeneity of urban environments, \acrshort{gnnui} incorporates diverse street-level features, such as infrastructure, points of interest, and weather. To more accurately model urban traffic distributions, we introduce a zero-inflated negative binomial distribution \citep{zhuang_uncertainty_2022}. Since interpolation targets are known in advance, the full graph, including nodes with missing values, is included during training, allowing the model to leverage their features and context from the outset. We share the full implementation of our approach in the accompanying repository for transparency and reproducibility \href{https://github.com/silkekaiser/GNNUI.git}{https://github.com/silkekaiser/GNNUI.git}, and to support it, we assemble and publish two large-scale, street-level urban traffic datasets: Strava cycling data from Berlin and New York City taxi data, each enriched with auxiliary urban features, offering complete spatio-temporal coverage across all street segments, and available for download \citep{kaiser_data_2025}. In addition, we empirically address two further questions: \textit{What is the most effective graph representation for interpolation?} and \textit{How well can traffic volumes be predicted under varying data scarcity?} To answer the first, we compare different adjacency matrix configurations. For the second, we evaluate model performance across a range of simulated sensor coverage levels, from sparse to near-complete. Most existing research focuses on settings with high sensor availability; however, real-world urban networks are characterized by extreme data sparsity, making performance under low coverage particularly important. 

In summary, our main contributions are the following:
\begin{itemize}
    \item We introduce \acrshort{gnnui}, a spatio-temporal GNN which interpolates citywide traffic volume while handling the unique challenges of urban contexts: infrastructural diversity through the inclusion of node features and handling of the skewed zero-inflated ground truth distribution through a zero-inflated negative binomial loss.
    \item We create and publish two large-scale, graph-based street-level traffic volume benchmark datasets, Strava cycling data from Berlin and taxi trip data from New York City, each enriched with a rich set of auxiliary features, and suitable for use as benchmark datasets for future research.
    \item We demonstrate that GNNUI outperforms baseline methods. We systematically evaluate its robustness across varying levels of simulated sensor coverage and demonstrate that it maintains strong performance, especially under extreme data scarcity. This is critical for real-world urban settings where sensor infrastructure is typically sparse, and reliable estimation is most needed.
\end{itemize}

\section{Related Work \label{sec:lit_review}}

Given the structural similarity between graphs and traffic networks, \acrshort{gnn}s have been widely adopted in transportation research. They are particularly effective at capturing spatial dependencies through graph structures and message passing \citep{xu_how_2019}. As a result, \acrshort{gnn}s are popular tools for traffic volume estimation, especially for short- and long-term forecasting across various transportation modes, including motorized traffic, urban rail systems, and ride-hailing platforms \citep{jiang_graph_2022}. While some studies applied GNNs to forecast urban data \citep{bai_a3t-gcn_2021, zhao_t-gcn_2020}, much of the existing research focused on forecasting tasks using benchmark datasets derived from highway sensor networks, such as PeMS and METR-LA \citep{yu_spatio-temporal_2018, yu_st-unet_2021, cai_traffic_2020, li_spatial-temporal_2021, bai_adaptive_2020}. This focus is unsurprising: highways and major arterial roads typically feature dense sensor coverage, and forecasting plays a critical role in real-time traffic management.

Despite the strong focus on forecasting, GNNs also have potential for interpolation tasks. Spatial interpolation (also known as spatial kriging or out-of-sample imputation) aims to estimate traffic conditions in areas without direct sensor coverage. Yet, these interpolation tasks received considerably less attention than forecasting in the literature. 
Traditional interpolation methods include kriging methods \citep{krige_statistical_1951, cressie_statistics_2015}, a pioneering approach using spatial correlation between data points and their observed labels. It assumes a Gaussian distribution, which is often not the case in real-world data. Although there are workarounds \citep{saito_geostatistical_2000}, circumventing this assumption is not always feasible. Other approaches harvested linear spatial dependencies to interpolate unobserved nodes, such as K nearest neighbor (KNN) \citep{cover_nearest_1967} or inverse distance weighting (IDW) \citep{lu_adaptive_2008}. Another approach was matrix/tensor completion \citep{bahadori_fast_2014, deng_latent_2016, zhou_kernelized_2012}. Crucially, often such methods are \emph{transductive}, meaning they cannot predict new nodes that have not been seen before; instead, one must retrain a model for each new graph structure. Other emerging methods, such as hybrid residual CNN–LSTM architectures \citep{ren_hybrid_2020}, versatile large language model-based frameworks \citep{li_urbangpt_2024, liu_spatial-temporal_2024, ren_tpllm_2024} or dynamic graph models \citep{huang_multi-view_2023}, have so far been applied only to traffic forecasting, not interpolation tasks.

Recent research demonstrated the ability of GNNs to perform spatial interpolation, by leveraging their \emph{inductive} nature to predict values for nodes not seen during training. Several approaches emerged, each addressing different aspects of the problem. 
\cite{appleby_kriging_2020} introduced kriging convolution networks (KCN), which leverage the strengths of kriging to use training labels as input and allow further features. KCN overlooks the temporal evolution of node labels in spatiotemporal datasets, necessitating retraining whenever new data is observed in sensor networks. While the authors demonstrated KCN across various applications, none involved traffic-related data.
In a similar vein, \cite{wu_spatial_2021} presented SATCN, a flexible framework for spatiotemporal kriging that leverages a spatial aggregation and temporal convolutional networks without requiring specific model design. The authors showcased the model's performance on METR-LA and other non-traffic-related data sets. 
They do not make use of data beyond the target measurements, and their spatial flexibility is only based on the distance between sensors.
Other models aim to refine the treatment of spatial and temporal dependencies. \cite{zheng_increase_2023} presented INCREASE, which focuses on encoding heterogeneous spatial relations and diverse temporal patterns. Their evaluation spanned multiple datasets, including METR-LA and urban motorized traffic data from Xiamen. The work relies in part on detailed flow data (e.g., transition probabilities) often unavailable in urban traffic settings, and whereas this work is one of the few to use additional features for spatial aggregation, it does not allow for the usage of temporally varying features. Additionally, while the model uses urban traffic data, it focuses exclusively on motorized vehicles. 
\cite{yao_spatiotemporal_2023} propose the GNN-STI model to address the challenge of interpolating sparsely and irregularly distributed ground truth data. Their evaluation includes both simulated and real-world datasets, demonstrating robust performance across varying levels of data scarcity. The proposed model is built on assumptions and spatial structures, such as diffusion-based processes in Euclidean space and Voronoi-adjacency, that are not well suited for traffic interpolation tasks in urban road networks, where data is constrained by network topology and directed flow. 
Building on inductive learning, \citep{wu_inductive_2021} proposed IGNNK, based on a graph convolutional network; it is an inductive model that allows, via a specialized masking algorithm, to interpolate traffic volumes for unseen locations using only the training labels of known locations. Their evaluation, conducted on METR-LA and PeMS-Bay, underscored the model’s applicability to traffic datasets. The paper has a well-suited masking strategy to allow the model to learn interpolation, but fails to take into account node features or a loss function suited for urban interpolation.

\section{Methods \label{sec:methods}}
\subsection{Model}
\begin{figure}[ht!]
    \centering    
    \includegraphics[width=0.6\textwidth]{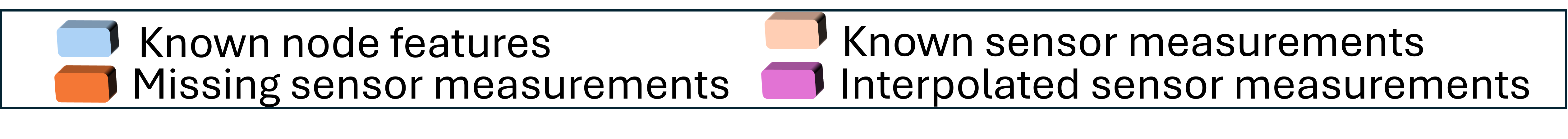}
    \vspace{0.5cm} 
        \begin{subfigure}[b]{0.49\textwidth}
        \centering
        \includegraphics[width=\textwidth]{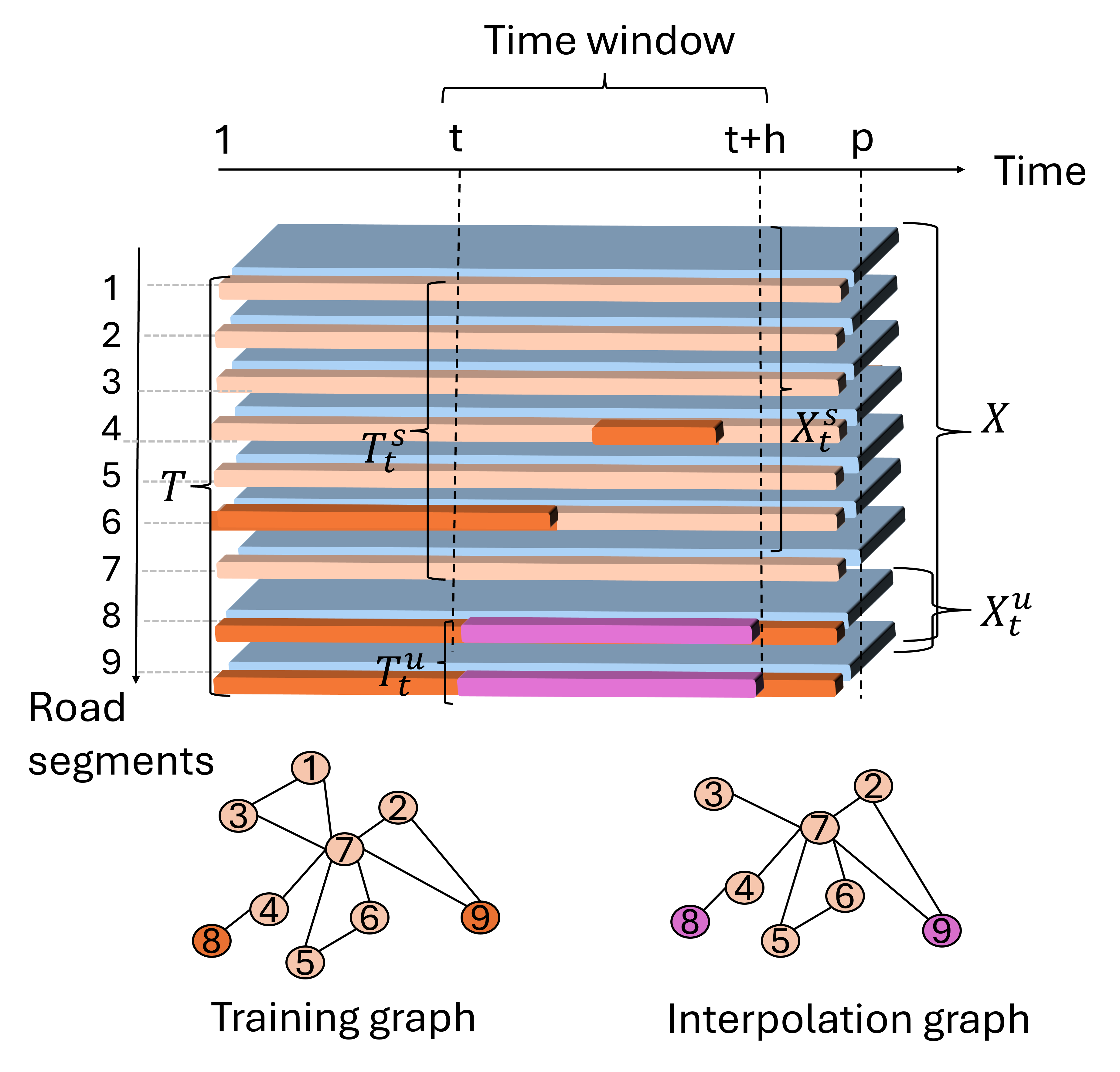}
        \caption{Using the entire graph in training.}
        \label{fig:explanaion_interpolation_a}
    \end{subfigure}
    \hfill
    \begin{subfigure}[b]{0.49\textwidth}
        \centering
        \includegraphics[width=\textwidth]{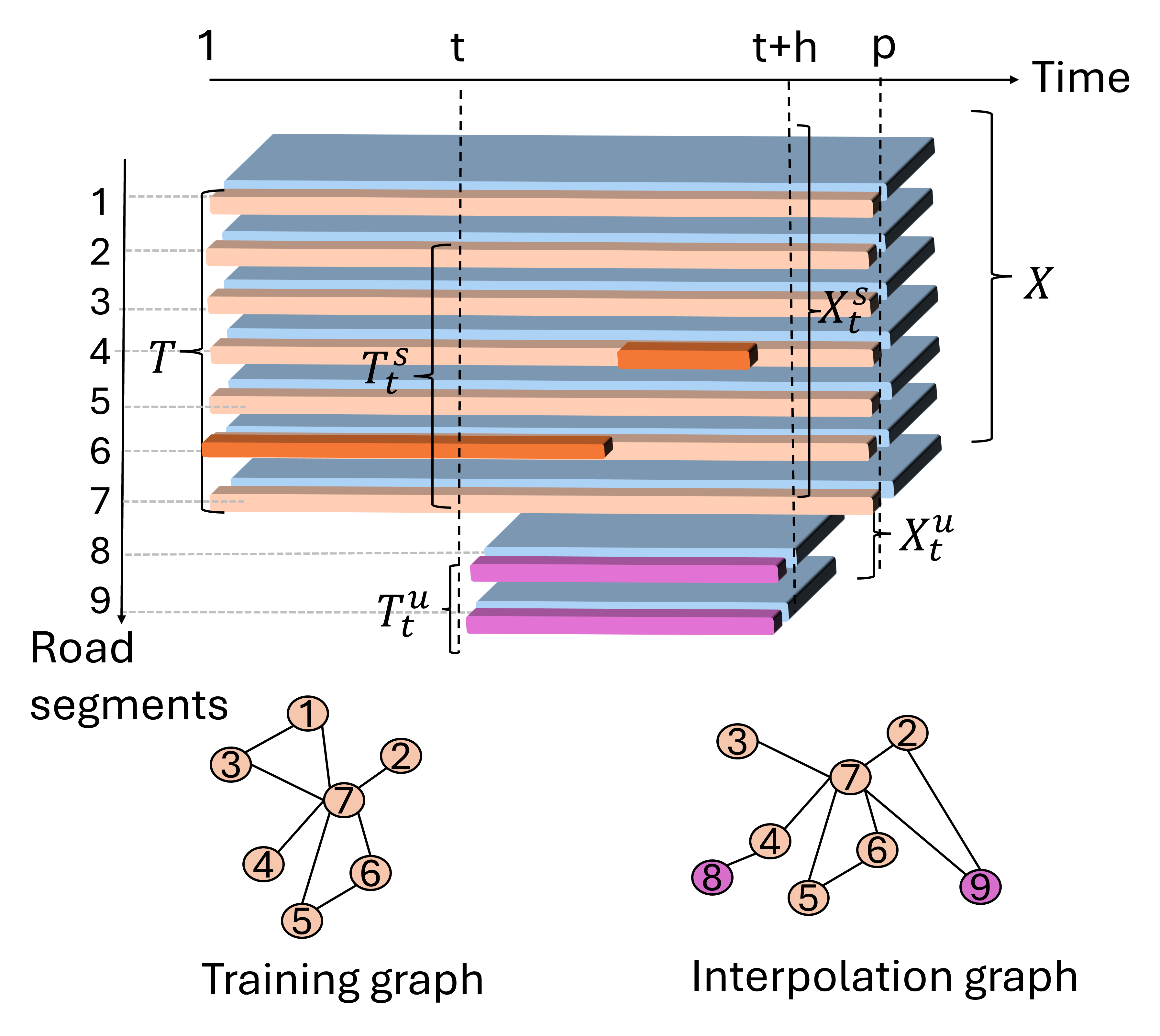}
        \caption{Without using the entire graph in training.}
        \label{fig:explanaion_interpolation_c}
    \end{subfigure}
    \caption{The figure illustrates how the interpolation is performed, whether the entire graph is leveraged in training. Across both figures, the seen locations are street segments $\{1,\cdots, 7\}$, and the unseen locations are street segments $\{8, 9\}$. Depicted are further the node feature training data $X$, the target variable training data $T$, the time window starting at $t$ of size $h$, as well as the node features of the unseen nodes $X_t^u$, the node features of the seen nodes $X_t^s$ and the corresponding target variables $T_t^u$ and $T_t^s$. }
    \label{fig:explanaion_interpolation}
\end{figure}

Throughout this section, we accompany our explanations with examples corresponding to Figure~\ref{fig:explanaion_interpolation_a}, representing our main specification. Differences related to Figure~\ref{fig:explanaion_interpolation_c} will be discussed at the end of the section.

\begin{figure}[!ht]
    \centering
    \includegraphics[width=0.55\linewidth]{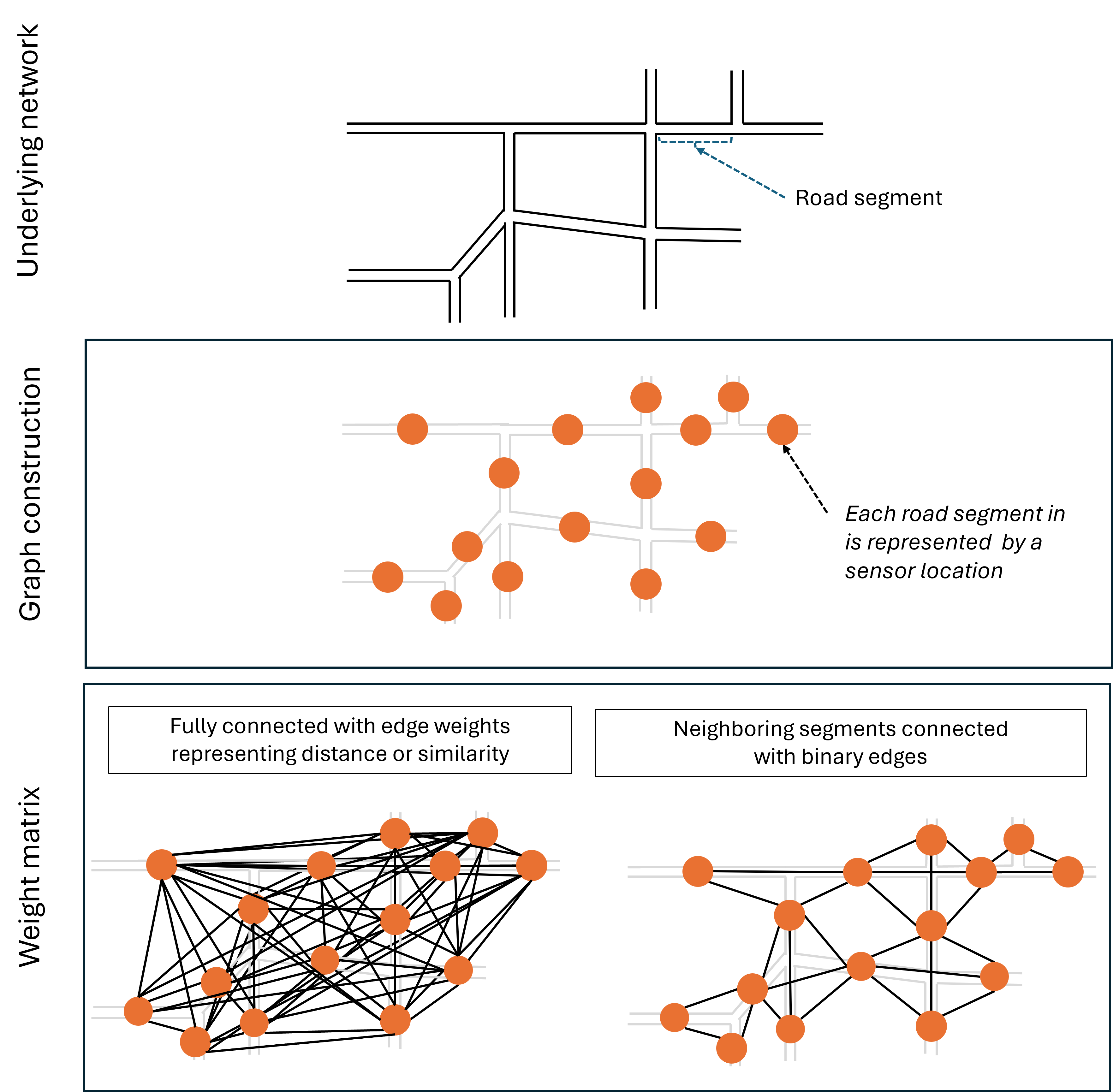}
    \caption{The graph and weight matrix are constructed based on the underlying street network, where each street segment is represented by a sensor location. The sensor locations form the nodes of the graph. Nodes are either fully connected (left), with edge weights representing distance or similarity (see Equations \ref{eq:graphweights_distance} and \ref{eq:graphweights_similarity}), or connected only to neighbouring nodes (right), using a binary adjacency matrix where weights are 1 for neighboring segments and 0 otherwise (see Equation \ref{eq:graphweights_binary}).}
    \label{fig:graph_explanation}
\end{figure}

\textbf{Network Design}. We model our network such that each street segment is thought to be equipped with a traffic volume sensor measuring the number of vehicles passing by (target variable). We use ``street segment'' interchangeably with ``sensor'', under the simplifying assumption of a one-to-one mapping. In practice, a street segment may have multiple sensors; however, since our framework treats all sensors on the same segment as recording the same traffic volume, extending our approach to handle multiple sensors per segment is straightforward. In our experiments, these sensors are simulated based on fully observed proxy datasets (taxi and Strava counts), allowing us to emulate different levels of sensing coverage. The nodes in our graph represent the $n$ sensor locations in the training data (Figure \ref{fig:graph_explanation}). The edges and their weights are determined by the adjacency matrix $W \in \mathbb{R}^{n\times n}$. The adjacency matrix can represent distance or similarity, resulting in a fully connected graph (lower left of the figure), or be binary, with edges only between direct neighbors (lower right). 

\textbf{Prediction Task}. The goal of our model is to interpolate the traffic volume for street segments that lack sensor data. For this, the model is provided with a) the target variables and node features of \textit{seen} (i.e., sensor-equipped), and
b) the node features of the \textit{unseen} (i.e., not equipped with sensors) segments.
All predictions are made over a fixed time window. The model predicts the target variable for unseen and seen nodes across the time window. \cite{wu_inductive_2021} compute the loss across all nodes, aiming to generalize the message-passing mechanism. Instead, we want our model to specialize in interpolating new locations. Thus, we compute the loss and the error metrics only on the interpolated nodes, while accounting for missing data (mathematical definition in \ref{appx:training_loss}). For example, in Figure \ref{fig:explanaion_interpolation_a}, the model uses the node features and the target variable of segments 1 to 7 (seen), to predict the target variable for segments 8 and 9 (unseen) on which the error metrics are computed.

\textbf{Node Features and Target Variables}. While most interpolation models rely primarily on the target variable from available sensors, we also incorporate node-level features. Features are important in urban contexts, where local characteristics significantly impact traffic volume. For example, a busy shopping street and a quiet residential street may be adjacent but have very different traffic volumes, differences that the model can only learn if features are included. All continuous input features, including the sensor measurements used as input to the model, are min-max scaled, and categorical features are one-hot encoded. No weighting is applied to any input features. We incorporate $k$ node features in our model. The data is available across $p$ time steps, the tensor $X \in \mathbb{R}^{n \times p \times k}$ contains the training node feature inputs for all sensor locations, while the target variable is stored in the tensor $T \in \mathbb{R}^{n \times p \times 1}$.  

\textbf{Time Window Approach to Model Temporal Dependency}.
To capture temporal dependencies, we adopt the fixed-length time window strategy from \cite{wu_inductive_2021}. For a given start time $t$, the model interpolates traffic volumes for the period $[t, t+h) = {t, t+1, \dots, t+h-1}$, where $h$ is the time window size. This assumes temporal correlation of traffic patterns within each window. 
While the employed time window approach effectively captures recurring traffic patterns (e.g., daily or weekly cycles), it is less suited to dynamic or abrupt changes in traffic, such as those caused by accidents, construction or other outstanding events. This limitation could be mitigated by incorporating real-time contextual signals (e.g., event or incident data) or by adopting adaptive temporal mechanisms that dynamically reweight recent observations. Exploring such adaptations remains an important direction for future work.

\textbf{Interpolation.}
In the interpolation (i.e., inference at test time) setting, denoted with subscript $t$, predictions are made for the time window  $h$. There are $n_t^s$ seen street segments (e.g., segments 1 through 7 in Figure \ref{fig:explanaion_interpolation_a}). For these segments, both the node features $X^s_t \in \mathbb{R}^{n_t^s \times h \times k}$ and the target values $T^s_t \in \mathbb{R}^{n_t^s \times h \times 1}$ are known. Conversely, there are $n_t^u$ unseen segments (e.g., segments 8 and 9). For these, only the node features $X^u_t \in \mathbb{R}^{n_t^u \times h \times k}$ are known, while the corresponding target variables $T^u_t \in \mathbb{R}^{n_t^u \times h \times 1}$ are missing. The goal of the model is to predict $T^u_t$ given $X^s_t$, $T^s_t$, and $X^u_t$, as illustrated in Figure \ref{fig:explanaion_interpolation_a}. 

\textbf{Training Through Masking Strategy}.
To train the model for interpolation, we adopt a masking strategy inspired by \cite{wu_inductive_2021}. We briefly summarize it here and provide full and technical details in \ref{appx:trainingthroughmasking}. Missing values are common in real-world sensor data and can be caused by malfunctions, maintenance, or construction activities. For instance, after removing outliers and accounting for gaps, 1.3\% of the Strava data and 1.0\% of the taxi data are missing. The masking strategy handles the missing data by replacing it with zero. During training, we simulate the interpolation process by constructing multiple training samples. A sample is a small training instance that includes a) a selected time window of length $h$, and b) a subset of nodes, randomly divided into supposedly seen and unseen nodes. Each sample mimics a localized spatio-temporal snapshot of the network, allowing the model to repeatedly learn the interpolation task from different perspectives. To mimic the real prediction setting, we mask the target variables of the supposedly unseen nodes by setting them to zero, just as we represent missing values in the test phase. The model is trained to predict these masked values using the target variables and node features of the supposedly seen and the node features of the supposedly unseen nodes. Loss is computed only on the masked nodes. Also different from \cite{wu_inductive_2021}, we ensure that time windows during training do not overlap, to make use of all training data and adopt a scheduled learning rate to improve convergence and stability.

Setting target variables to zero can be problematic because the model cannot distinguish between genuinely zero values, those set to zero due to masking, and those set to zero because they are missing. To resolve this ambiguity, and differing from the original masking strategy in \cite{wu_inductive_2021}, we include two binary indicators in each node’s feature vector to denote masking and missingness. These are treated as standard input features and are already accounted for in the feature vector dimensionality k (see \ref{appx:trainingthroughmasking} for a mathematical formulation).

\textbf{Entire Graph Utilization}. In our default model specification, we train on the entire graph, including all sensor locations (e.g., nodes 1 through 9 in Figure \ref{fig:explanaion_interpolation_a}). The target variables of the unseen nodes (e.g., segments 8 and 9) are never used during training, they are treated as missing throughout. However, their node features are included, allowing the model to learn from their structural and contextual information without accessing their true labels. Training on the entire graph is a modeling choice, not a necessity. Since our model is inductive, it can also be trained without access to the unseen nodes (Figure \ref{fig:explanaion_interpolation_c}).

\textbf{Test Data from Same Time as Training Data}. The test data comes from the same time period as the training data (in both Figure \ref{fig:explanaion_interpolation_a} and \ref{fig:explanaion_interpolation_c}). We train the model on time steps $1$ through $p$. The test is then performed on segments 8 and 9 during the time period $1$ through $p$. Since the test time overlaps with training, the model has already seen parts of the input (i.e., $X_t^s$, $X_t^u$, and $T_t^{s}$ under consideration of truly missing values). While this setup focuses on spatial interpolation, it requires retraining the model if predictions are needed for a different time. This is particularly suited for retrospective urban planning applications, where historical traffic distributions inform infrastructure decisions. In contrast, an alternative setup could consist of the model generalizing to both unseen locations and future times, meaning that no test inputs are observed during training. This is valuable in applications requiring continuously updated traffic forecasts, such as dynamic traffic control. This is explained and parallel to the results in this study, evaluated in an ablation study in \ref{appx:temporal_generalization}.

\subsection{Adjacency matrices}
Graph Neural Networks work with graph-structured data and, therefore, greatly depend on how the graph is defined \citep{ye_how_2022}. Various approaches exist for constructing adjacency matrices, including fixed, dynamic, or attention-based methods (i.e., graph attention networks). Fixed adjacency matrices are widely used, as they provide an interpretable and consistent neighborhood structure, particularly useful when modeling traffic systems with relatively stable spatial layouts, and are computationally efficient. However, fixed adjacency matrices fail to capture evolving spatial dependencies. Alternative strategies, such as learning dynamic adjacency matrices during training \citep[e.g.,][]{zhang_spatio-temporal_2025}, may better capture evolving traffic conditions, but at a higher computational cost. Similarly, graph attention networks \citep[e.g.,][]{zheng_gman_2020} offer another alternative by learning neighbor relevance directly from data. These models can be computationally efficient at scale, making them promising for large graphs, though they come with added architectural complexity. Given these trade-offs, we focus on fixed adjacency matrices in this work. Specifically, we investigate three types of fixed adjacency matrices commonly used in traffic modeling: binary, distance-based, and similarity-based. These were selected for their ease of construction and theoretical grounding in spatial and functional relationships between traffic sensors.

Binary adjacency matrices capture the local topology of the urban street network by connecting nodes (i.e., street segments) that are directly linked in space. They are widely used in the traffic domain due to their simplicity and ability to preserve physical road connectivity \citep{guo_attention_2019, ye_how_2022, zhao_t-gcn_2020}. A binary adjacency matrix is defined using street segments $v_i$ and $v_j$ as: \begin{equation}
    w_{ij}=\begin{cases}
    1, & \text{if }v_i \text{ and } v_j \text{ meet at an intersection},\\
    0, & \text{otherwise}.
  \end{cases}
  \label{eq:graphweights_binary}
\end{equation}

Distance-based adjacency matrices extend this notion by weighting all node pairs based on spatial proximity, thus producing fully connected graphs where nearby nodes exert stronger influence and are commonly used in the traffic domain\citep{li_diffusion_2018, wang_origin-destination_2019}. Following previous work \citep{li_diffusion_2018}, we define the distance-based adjacency matrix as:
\begin{equation}
    w_{ij}=\exp \left(-\left(\frac{\text{dist}(v_i, v_j)}\sigma\right)^2\right),
    \label{eq:graphweights_distance}
\end{equation}
where $\sigma$ is a normalization factor. We compute $\text{dist}(v_i, v_j)$ using (a) birdfly distance, (b) street distance, and (c) travel time, with the last two derived from OpenStreetMaps.

One limitation of distance-based and binary matrices is that they overlook nodes that, while geographically distant, exhibit similar traffic patterns due to their functional similarity \citep{jiang_graph_2022, geng_spatiotemporal_2019}. Thus, we explore similarity-based adjacency matrices, in which nodes are connected with a higher weight if they exhibit functionality similarity \citep{jiang_graph_2022, ye_how_2022}. Inspired by prior work, we compute our similarity-based adjacency matrix based on \acrlong{poi} (\acrshort{poi}) and built infrastructure \citep{geng_spatiotemporal_2019, ge_global_2020},  using a cosine similarity \citep{shi_predicting_2020, he_dynamic_2020}. 
Let $X^{infra} \in \mathbb{R}^{n \times k'}$ denote the matrix of time-invariant infrastructure features, where $k'$ is the number of such features. These form a subset of the full set of node features, so $k' \leq k$. The adjacency matrix is:
\begin{equation}
    w_{ij}=\cos{(X_i^{infra}, X_j^{infra})}= \frac{X_i^{infra} \cdot X_j^{infra}}{||X_i^{infra}|| \cdot ||X_j^{infra}||}
    \label{eq:graphweights_similarity}
\end{equation}

\subsection{Model pipeline and loss}
The neural network consists of three Diffusion Graph Convolutional Network (DGCN) layers, similar to the work of \cite{wu_inductive_2021} and \cite{zhuang_uncertainty_2022} (mathematical definition of all layers and pipeline depiction in \ref{appx:model_pipeline}), and an output layer. This output layer is tailored to the chosen loss function and outputs the parameters needed to compute the respective loss. In our datasets, traffic volume is characterized by a right-skewed distribution with an inflated number of zeros (Figure \ref{fig:network_comparison_combined}). The training loss should account for this. Mean square error (MSE) losses, commonly used in highway models, are ill-suited for this setting. They optimize for the mean and are overly sensitive to large values. This can cause the model to overestimate lower traffic volumes and underestimate higher ones. To address this, we experiment with different loss functions: we use the more robust \acrlong{mae} (\acrshort{mae}) and Gaussian Negative Log-Likelihood (GNLL), as well as the \acrlong{nb} (\acrshort{nb}) and \acrlong{zinb} (\acrshort{zinb}) losses from \cite{zhuang_uncertainty_2022}. These are tailored to discrete, skewed data, with \acrshort{zinb} additionally handling excess zeros well (mathematical definition of losses in \ref{appx:zinb_and_nb}).
 
\subsection{Evaluation}
\textbf{Training, validation, and test set}. We hypothesize that the amount of ground truth data, i.e.~the share of street segments equipped with a sensor, affects the model's predictive performance. To simulate different levels of data scarcity, we fix 10\% of street segments for testing and vary the share of street segments available for training and validation. We start with all 90\% of segments (split 9:1 for training and validation) and gradually reduce the available share to 1\%. All time steps of a given segment belong to the same subset. No temporal split is used because the task is spatial interpolation rather than temporal forecasting. 

\textbf{Error metrics}. We assess interpolation performance using the \acrshort{mae} and \acrlong{rmse} (\acrshort{rmse}) ( \ref{apx:errors}). We avoid percentage-based metrics due to the prevalence of zeros and outliers in the data.
Due to the skewed and zero-inflated nature of the data, we additionally report the \acrlong{kl} (\acrshort{kl}) to compare predicted and actual distributions, as done in related work on similar data \citep{zhuang_uncertainty_2022}. Since \acrshort{kl} quantifies the disparity between two distributions, lower values are preferable. To compute the KL divergence, we require the predicted and actual probability distributions, which we approximate by discretizing the continuous values into bins. To evaluate how well the model captures the distribution’s zero-inflated nature, we compute the true-zero rate. Which measures the proportion of true zeros correctly predicted as (near-)zero values, with higher values being desirable (mathematical definition in \ref{appx:true_zero_rate}).

\textbf{Baselines}. We compare against two baselines: IGNNK \citep{wu_inductive_2021}, a graph-based model for interpolation that shares architectural elements with ours (trained here with a binary adjacency matrix, a validation set, and an MSE loss as in the original implementation), and XGBoost, a strong non-graph-based baseline for feature-rich traffic data, which has previously shown good performance in interpolation tasks \citep{kaiser_counting_2025}, trained with MAE.

\section{Data \label{sec:data}}
To evaluate our model, we require urban datasets with traffic volume information at the street-segment level, combined with rich auxiliary features. However, to the best of our knowledge, no city provides complete sensor-based traffic volume data across its entire street network. Therefore, we assembled two large-scale datasets ourselves: Berlin Strava and New York City taxi data. For both, we simulate full sensor coverage by aggregating trajectory data to the street segment level and enriching them with extensive contextual features. While we refer to sensors throughout, all evaluations are based on these proxy datasets, not actual sensor data. While Berlin data reflects bicycle traffic, NYC data captures motorized mobility, together offering a diverse and representative testbed for urban interpolation tasks.

\subsection{Berlin Strava and New York taxi volume measurements}
The Strava dataset contains the daily number of bike trips recorded via the Strava app \citep{strava_metro_strava_2024} at the street segment level in Berlin from 2019 to 2023. Strava aggregates publicly shared trips across sidewalks, bike paths, and vehicle lanes, rounding volumes to the nearest multiple of five for privacy. We further aggregate these measurements to the street segment level for our analysis. Although the data is biased toward young, male, and sporty users (descriptive statistics in \ref{apx:strava_descriptives}), it still provides valuable insights into urban cycling patterns \citep{kaiser_counting_2025, lee_strava_2021}.

The taxi dataset covers hourly taxi trips in Manhattan for January and February 2016, based on publicly available data  \citep{new_york_taxi_and_limousine_commission_new_2016}. For computational feasibility, we focus on trips within Manhattan during this period, which includes GPS coordinates for pickup and drop-off locations. From these, we reconstruct trajectories using the Open Source Routing Machine, which computes vehicle routes based on street network data, including geometry, one-way streets, and turn restrictions \citep{luxen_real-time_2011}. Based on these trajectories, we compute the number of taxis passing per street segment \citep{boeing_modeling_2025}.

For both datasets, we remove observations deviating more than three standard deviations from a segment's mean. Due to the strong right skew, this only removes high values. These values are caused by events like exception occasions temporarily disrupting normal traffic patterns, such as bike demonstrations or street protests. Since our focus is to model normal traffic behavior, e.g., to inform infrastructure planning, we prioritize capturing average conditions over rare anomalies. Filtering per segment preserves meaningful variation, such as consistently busy streets, while improving overall data quality. Even after filtering, the distributions remain highly right-skewed  (Figure \ref{fig:network_comparison_combined}).

\subsection{Graph structure}
We base the Berlin graph on the city's official cycling network \citep{senate_department_for_the_environment_mobility_consumer_and_climate_protection_berlin_radverkehrsnetz_2024}, which includes both major traffic routes and smaller streets with high cycling volumes (Figure \ref{fig:network_comparison_berlin}). This network, smaller than Berlin’s full street network, makes the task computationally manageable and comprises 4,958 street segments. For New York, we construct the graph using all public streets in Manhattan obtained from OpenStreetMap data, resulting in a network of 8,156 street segments (Figure \ref{fig:network_comparison_ny}).

\subsection{Predictive features}
We compile further features that have been helpful in previous research in predicting cycling volume and motorized traffic, which also contribute in our model (see \ref{apx:feature_importance} for a feature importance analysis). We outline this related data below, and include a list of all features in \ref{apx:list_of_node_features}. 

For both Strava and taxi data, we collect data on built infrastructure and \acrshort{poi}.
Built infrastructure influences cycling behavior \citep{yang_towards_2019}; for example, street design and dedicated cycling facilities affect both actual and perceived safety, which in turn shape bicycle usage patterns \citep{costa_unraveling_2024, moller_cyclists_2008}. \acrshort{poi}s are predictors of taxi and bicycle demand \citep{askari_taxi_2020, fazio_bike_2021,strauss_spatial_2013}; streets with more shops, for instance, tend to attract higher traffic. We obtain infrastructure data from \cite{openstreetmap_contributors_planet_2017} for every street segment, including the speed limits, surface type, and road classification. For Strava, we add measures of cyclability and cycle lane types; for the taxi data, we include the number of lanes and the presence of parking.  \acrshort{poi} features include the number of shops, hotels, industrial facilities, and hospitals within 50m, 100m, 250m, and 500m of each street segment. We also compute the distance of every segment to the city center. For the taxi data, we also record the number of educational facilities, restaurants, bars, cafés, railway stations, bus stops, and long-distance bus stops. For the Strava data, we were further able to obtain data on the presence of parks, forests, cemeteries, and other similar features \citep{berlin_open_data_nutzung_2022,  senate_department_for_urban_development_building_and_housing_lebensweltlich_2023}. The city of Berlin provides these features for planning areas of around 2km$^2$ in size. The city collected the data once in 2015; thus, while there is spatial variation, there is no temporal variation. 

We also include network connectivity measures, which reflect how integrated a street segment is within the overall network. Connectivity influences cycling and motorized traffic demand by shaping movement patterns through the network’s geometric properties \citep{hillier_network_2005, hochmair_estimating_2019, schon_scoping_2024}. These measures, computed based on the designated graph structure, include degree, betweenness, closeness, and clustering coefficient (mathematical definitions in \ref{apx:connectivity_measures}). 

Also, weather influences both motorized traffic \citep{koesdwiady_improving_2016} and cyclists' decisions to ride \citep{miranda-moreno_weather_2011}. We include daily weather features for Berlin and daily and hourly weather features for New York \citep{meteostat_weathers_2022}, e.g., covering sunshine duration, precipitation, and temperature.

We include temporal indicators for both the Strava and taxi datasets, such as the day of the week, weekends, month, and public or school holidays \citep{senate_department_for_education_youth_and_family_ferientermine_2023, new_york_city_department_of_education_2011-12_2017}. For the taxi data, we also add the hour of the day, and for the Strava data, the year.

For the Strava data, we also include features on motorized traffic and socioeconomic indicators. Motorized traffic is a strong predictor of cycling volume \citep{kaiser_counting_2025}, likely due to shared temporal patterns such as commuting. Given limited sensor coverage, we use the average vehicle speed and volume within a 6km radius of each segment, the smallest feasible radius ensuring data availability, and add citywide traffic indicators \citep{berlin_open_data_verkehrsdetektion_2024}. Socioeconomic factors like age and gender also influence cycling behavior \citep{goel_cycling_2022}. We obtain annual socioeconomic data at the planning area level \citep{berlin-brandenburg_office_of_statistics_kommunalatlas_2023}, including gender distribution, age distribution, and population density, using year-specific data for 2019–2020 and the latest available data (from 2020) for 2021–2023.

\section{Results and Discussion \label{sec:results}}

\subsection{Performance of GNNUI \label{sec:results_performance}}
We conduct an ablation study of GNNUI by cumulatively removing individual components to assess their contribution to improving the performance in the urban traffic volume interpolation tasks. These components include: (i) using the entire graph during training, (ii) incorporating auxiliary node features, (iii) computing loss only on masked nodes, (iv) adding indicators for masked and missing values, and (v) using \acrshort{nb} and \acrshort{zinb} losses to better capture skewed, zero-inflated distributions. All components are independently applicable, except for (iv), which depends on the inclusion of node features (ii). We use a data-rich scenario for the ablation (90\% of segments for training/validation), as most related studies work with data-rich settings. We compute the ablation study, using \acrshort{gnll}, which is a standard negative log-likelihood loss and serves as a baseline before exploring more tailored losses, and \acrshort{mae} as a widely used performance metric, choosing MAE over RMSE due to its robustness to outliers, which is particularly important given the right-skewed nature of our data.
\begin{table}
\centering
    \caption{Evaluation metrics for ablation study with Strava Berlin and taxi New York datasets. N/A indicates that the model does not produce a probabilistic output required to compute NLL.}
    \label{Table:table1_interpolation_appx}
    \tiny
        \begin{tabular}{lccccc|rrrrr|rrrrr}
            \toprule
            \multicolumn{1}{l}{}&  \multicolumn{5}{c}{Components}& \multicolumn{5}{c}{Strava}&   \multicolumn{5}{c}{Taxi}\\
            \midrule
            & \rotatebox{90}{Entire graph (i) }& \rotatebox{90}{Node features (ii)} & \rotatebox{90}{Loss on masked only (iii)}   &\rotatebox{90}{Feature missing/masked (iv)} &\rotatebox{90}{training loss (v)}&     &  &  &  & &  &  &  &    \\
            &     &  &   &   &    &MAE  &RMSE  & KL & Zero &NLL &MAE  &RMSE  & KL & Zero &NLL \\
            \midrule
IGNNK &   &   &   &   & MSE & 9.84 & 17.5 & 0.18 & 0.096 & N/A & 35.83 & 78.8 & 0.08 & 0.386 & N/A\\
XGBoost &   &   &   &   & MAE & 8.20 & 16.2 & 0.38 & 0.542 & N/A & 24.62 & 66.7 & 0.17 & 0.652 & N/A \\  
            \cline{2-16}
GNNUI & \checkmark & \checkmark & \checkmark & \checkmark & ZINB & \textbf{7.14} & \textbf{13.8} & \textbf{0.10} & 0.603 & 0.636\textsuperscript{*} & 23.03 & 56.8 & \textbf{0.08} & 0.731 & 0.638\textsuperscript{*}  \\
  & \checkmark & \checkmark & \checkmark & \checkmark & NB & 7.52 & 13.9 & 0.10 & 0.474 & 0.245\textsuperscript{*} & 27.11 & 63.4 & 0.23 & 0.582 & 0.342\textsuperscript{*}  \\
              \cline{2-16}
  & \checkmark & \checkmark & \checkmark & \checkmark & GNLL & 8.18 & 14.5 & 0.70 & 0.236 & 221 & 19.91 & \textbf{52.7} & 0.44 & 0.275 & 4537 \\
  & \checkmark & \checkmark & \checkmark &   & GNLL & 8.05 & 14.8 & 0.83 & 0.244 & 234& 23.26 & 62.5 & 0.52 & 0.276 & 6353   \\
  & \checkmark & \checkmark &   &   & GNLL & 7.80 & 16.3 & 1.19 & 0.333 & 283 & 25.40 & 74.1 & 0.85 & 0.259 & 8837 \\
  & \checkmark &   &   &   & GNLL & 9.46 & 17.0 & 1.16 & 0.057 & 303   & 35.15 & 77.3 & 0.73 & 0.000 & 9616 \\
  &   &   &   &   & GNLL & 9.24 & 16.8 & 1.04 & 0.042 & 296 & 32.55 & 73.3 & 0.54 & 0.019 & 8627 \\

            \cline{2-16}

  & \checkmark & \checkmark & \checkmark & \checkmark & MAE & 7.21 & 14.1 & 0.25 & \textbf{0.621} & N/A & \textbf{18.81} & 56.9 & 0.36 & \textbf{0.736} & N/A \\ 
  & \checkmark & \checkmark & \checkmark &   & MAE & 7.81 & 15.3 & 0.55 & 0.450 & N/A & 21.97 & 67.6 & 0.63 & 0.731 & N/A \\
  & \checkmark & \checkmark &   &   & MAE & 8.16 & 16.4 & 0.94 & 0.327 & N/A & 30.09 & 90.3 & 2.37 & 0.332 & N/A  \\
  & \checkmark &   &   &   & MAE & 9.65 & 19.8 & 2.08 & 0.381 & N/A& 33.48 & 95.9 & 3.47 & 0.671 & N/A \\
  &   &   &   &   & MAE & 9.69 & 20.0 & 2.13 & 0.438 & N/A & 33.62 & 96.1 & 3.45 & 0.698 & N/A \\
  \bottomrule
    \end{tabular}
        \vspace{1ex}
    \begin{minipage}{\textwidth}
    \tiny
    \centering \textsuperscript{*}NLL values from GNLL, ZINB, and NB are not directly comparable, as they are computed over different types of probability distributions.
    \end{minipage}
    
\end{table}

Results are shown in Table~\ref{Table:table1_interpolation_appx}. Overall, GNNUI performs best when trained with the ZINB loss. This configuration yields the lowest \acrshort{kl} across all settings and maintains high true zero rates (0.603 and 0.731). While the MAE loss results in slightly higher true zero rates for the taxi data (0.621 and 0.750), this comes at the expense of increased \acrshort{kl}, indicating a poorer fit to the overall distribution. If we apply a stricter threshold for what qualifies as a true zero (i.e., reducing the threshold $\tau$ from 0.99), ZINB outperforms MAE even on the true zero metric, further supporting its robustness for zero-inflated data. Dataset-specific results also highlight important differences. For the Strava data, both MAE and RMSE are minimized using the fully specified model with the ZINB loss. For the taxi data, however, the best performance is observed with models trained using MAE or GNLL losses (MAE of 23.03 vs. 18.81 with MAE loss and RMSE of 56.8 vs. 52.7 with GNLL loss). We hypothesize that this stems from differing data characteristics: Strava contains a higher share of zero observations (47.7\%) compared to taxi data (24.4\%), which plays to ZINB’s strength in capturing zero-value observations. In contrast, GNLL and MAE appear better suited to modeling the long-tail distribution of high-volume traffic, albeit with trade-offs in KL divergence and true zero preservation. 

Beyond error metrics, the ablation study reveals that each architectural component contributes positively to model performance. Notably, incorporating node features consistently enhances results across both datasets, underscoring the value of auxiliary data in urban environments. For example, on the Strava dataset, adding node features to a GNLL-trained model reduces MAE from 9.46 to 7.80 and RMSE from 17.0 to 16.3, while the KL divergence remains stable (1.16 to 1.19) and the true zero rate increases substantially from 0.057 to 0.333. While the magnitude of improvements varies, the relative benefits of components like training on masked nodes and including indicators for missing/masked data remain stable across almost all specifications, suggesting that these modeling choices are robust across different contexts. For the Taxi dataset, training on masked nodes with the MAE loss improves performance significantly: MAE decreases from 30.09 to 21.97, RMSE from 90.3 to 67.6, KL divergence from 2.37 to 0.63, and the true zero rate increases from 0.332 to 0.731. However, these findings are not absolute. Model configurations may need adjustment based on specific data characteristics. For instance, in datasets aggregated to coarser temporal resolutions (e.g., yearly rather than daily), the zero-inflation may diminish, making a negative binomial loss (NB) more appropriate. Conversely, if modeling is limited to denser urban cores where traffic volumes follow more symmetric distributions, losses like MAE might be a better fit.

In addition to the interpolation task, we evaluate GNNUI in a more challenging setting where it must predict values at new locations during future time periods. This setup allows us to assess its temporal generalisation capabilities and compare them to the literature focusing on forecasting. Although GNNUI was not specifically designed for forecasting, it still outperforms the standard baselines in this more challenging environment, demonstrating its robustness and flexibility across different tasks. See \ref{appx:temporal_generalization} for details.

While Strava and taxi data represent biased subsets of overall traffic, they still capture meaningful spatiotemporal patterns typical of broader mobility behavior. GNNUI's strong performance across both datasets demonstrates its ability to learn from such patterns under real-world, noisy conditions. This suggests that GNNUI can generalize well to other types of traffic data, including more representative datasets such as general motorized traffic or pedestrian flows.
 
In summary, the full model with the ZINB loss offers the best overall performance, achieving the lowest \acrshort{kl}, and competitive true zero, MAE, and RMSE values. Its ability to capture the true distribution is key for informed urban planning. Reliably distinguishing between high- and low-traffic streets informs decisions around infrastructure investment and policy prioritization.


\subsection{Different Adjacency Matrices \label{sec:results:different_adjacency_matrices}}
We evaluate \acrshort{gnnui} using binary, distance-, and similarity-based adjacency matrices to identify the most effective network structure for urban traffic interpolation (Table \ref{tab:compare_adjacency_matrices}). The analysis uses GNNUI with ZINB loss, which yields the best overall results.

We find that the choice of the adjacency matrix significantly affects performance. The binary matrix consistently performs best.  It achieves the lowest MAE, 7.14 for Berlin (vs. 8.92–9.73) and 23.03 for New York (vs. 30.99–40.50), as well as the lowest RMSE, 13.8 (vs. 16.8–17.4) and 56.8 (vs. 70.9–97.9), respectively. It also delivers the lowest KL divergence, 0.10 (vs. 0.12–0.21), and the highest true-zero rates: 0.603 (vs. 0.461–0.403) and 0.731 (vs. 0.516–0.006). We also evaluate hybrid adjacency setups, where multiple matrices are combined by training separate subnetworks for each. While this increases computational cost, it does not lead to meaningful performance gains, likely because the added edges introduce noise or less informative relationships, weakening the strong local signal captured by the binary adjacency. Full results and technical details are reported in \ref{appx:different_adjacency_matrices}. 

The performance advantage of binary adjacency matrices likely arises because they restrict message passing to immediate neighbors. In contrast, distance- and similarity-based matrices introduce weaker or noisier connections, diluting the relevance of the aggregated information. By focusing on clear, local dependencies, binary matrices provide a sharper and more informative signal for interpolation. Additionally, binary adjacency matrices offer clear computational advantages. Their simplicity reduces training time, making them a practical and robust choice for city-scale applications. 

It should be noted that this performance advantage is demonstrated on Strava and taxi data, which represent biased subsets of total urban traffic. Nonetheless, the strong and consistent results across both datasets suggest that the relative effectiveness of different types of adjacency matrices is likely robust and transferable to other traffic domains.

\begin{table}[h!]
\centering
\small
    \caption{Comparison of GNNUI given different adjacency matrices.} %
  \label{tab:compare_adjacency_matrices}%
    \begin{tabular}{l|rrrr|rrrr}
        \toprule
        \multicolumn{1}{c}{Adjacency matrix}&      \multicolumn{4}{c}{Strava Berlin}& \multicolumn{4}{c}{Taxi New York} \\
        \midrule
        &          MAE &RMSE & KL &  Zero &  MAE &RMSE & KL & Zero  \\
        \midrule
            Binary &    \textbf{7.14} & \textbf{13.8} & \textbf{0.10} & \textbf{0.603} &    \textbf{23.03} & \textbf{56.8} & \textbf{0.08} & \textbf{0.731} \\
            Similarity &    8.92 & 17.4 & 0.21 & 0.434 &    40.50 & 97.9 & 1.92 & 0.006 \\
            Distance Birdfly    & 9.47 & 17.2 & 0.18 & 0.403    & 32.52 & 70.9 & 0.28 & 0.516 \\
            Distance Street &    9.73 & 17.4 & 0.16 & 0.417 &    30.99 & 73.4 & 0.51 & 0.506 \\
            Distance Time &    9.20 & 16.8 & 0.12 & 0.461 &    31.68 & 74.1 & 0.52 & 0.511 \\
        \bottomrule
        \end{tabular}
\end{table}

\subsection{Varying data scarcity during training data}
\begin{figure}[h]
    \centering
    \begin{subfigure}{0.49\linewidth}
        \includegraphics[width=\linewidth]{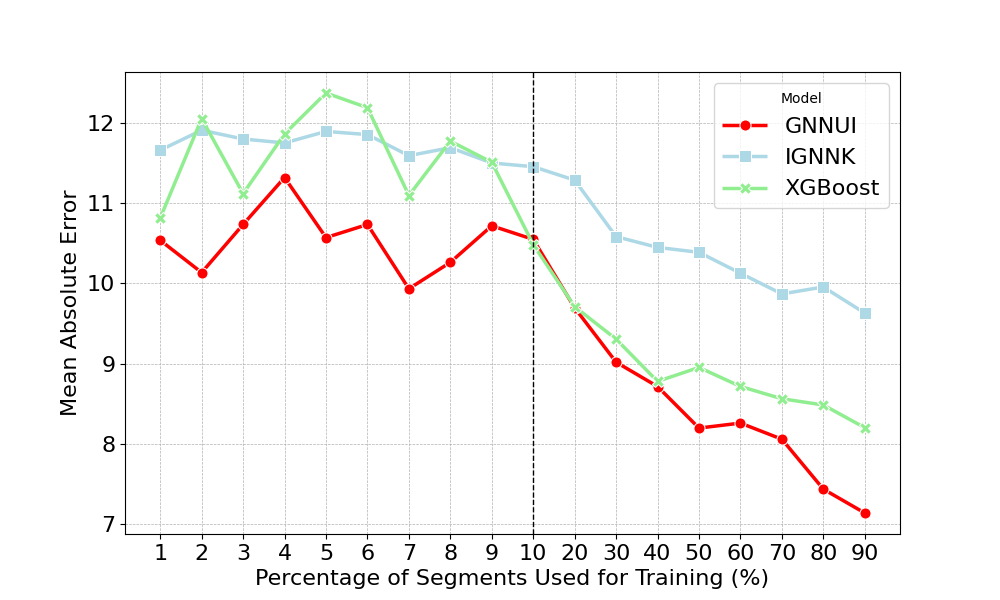}
        \caption{Strava}
        \label{fig:percentage_performance_Strava}
    \end{subfigure}
    \hfill
    \begin{subfigure}{0.49\linewidth}
        \includegraphics[width=\linewidth]{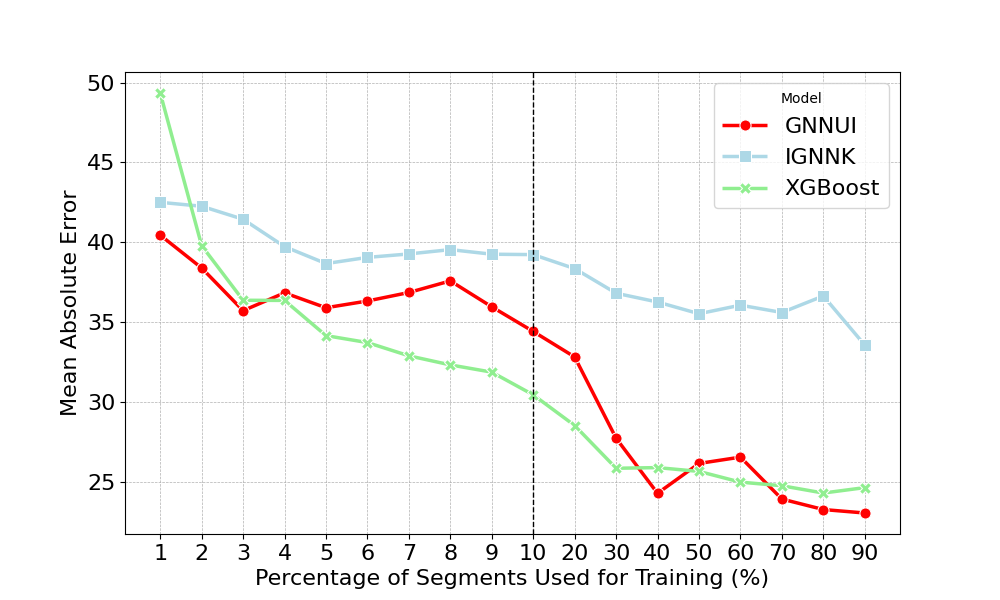}
        \caption{Taxi}
        \label{fig:percentage_performance_Taxi}
    \end{subfigure}
    \caption{Performance of GNNUI, IGNNK, and XGBoost under varying levels of data scarcity. Notice that there is a structural break on the x-axis at 10\%.}
    \label{fig:percentage_performance_main}
\end{figure}

Data scarcity poses a challenge for the interpolation of urban traffic volumes. We investigate how GNNUI's performance changes with decreasing sensor availability. This analysis uses a binary adjacency matrix.
We evaluate GNNUI across sensor coverage levels from 1\% to 90\%—using a step size of 1 percentage point until 10\%, and 10 percentage points thereafter. To reduce variability across experiments, we sample from the available ground truth sensor locations in a nested fashion: lower-coverage training sets are subsets of higher-coverage ones (e.g., the 70\% set is fully contained within the 80\% and 90\% sets). As before, we include the entire graph during training: all node features are visible to the model, but the target variables of nodes excluded from the training set are treated as missing. That is, nodes not selected for training still appear in the graph with their features, but their true traffic volumes are masked and not used for loss computation. The test set remains fixed, using the same 10\% of nodes for evaluation across all experiments. Results are shown in Figure~\ref{fig:percentage_performance_main}. For completeness, we report results for GNNUI using distance and similarity-based adjacency matrices in \ref{appx:percentages}. Both datasets reflect specific mobility subpopulations rather than the full spectrum of urban traffic. The consistent findings across these two distinct cases, however, suggest that the learnings are transferable to other traffic contexts.

We find that GNNUI achieves comparable performance to the baselines while requiring significantly less training data. For example, XGBoost achieves its best MAE of approximately 8.2 on the Strava data, but only with 90\% sensor coverage. In contrast, GNNUI reaches a similar MAE with just 50\% coverage. We observe a similar pattern for the taxi data, where GNNUI with only 40\% of the training data matches the performance of XGBoost trained on 80\% coverage. 

Also, we observe that GNNUI outperforms both IGNNK and XGBoost across a wide range of sensor coverage levels. For the Strava data, GNNUI outperforms the baselines across almost all coverage levels, from extremely sparse to dense. For the taxi data, GNNUI performs particularly well at both very low ($\leq$ 3\%) and high ($\geq$ 70\%) sensor coverage levels. GNNUI’s robustness under data scarcity is a critical strength for real-world urban applications, where sensor infrastructure is often extremely limited: in Berlin, bike sensor coverage corresponds to approximately 0.8\% of the cycling network, and in New York City, motorized traffic sensor coverage reaches only around 0.3\% of street segments.
GNNUI’s stronger and more consistent advantage on the Strava data stems from two main factors. The higher proportion of zero counts in Strava makes the ZINB loss particularly effective. And cycling traffic is more tightly linked to the street network structure, which graph-based models like GNNUI capture well. 
   
Lastly, we find that performance degradation at low sensor coverage levels is not severe, indicating that GNNUI can still produce reliable estimates even with little training data. For Strava, \acrshort{gnnui} achieves an MAE of 10.5 at 1\% coverage, improving to 7.1 at 90\%. For the taxi data, the MAE drops from 40.4 to 23.0 in the same range.  Improvements occur gradually across the coverage spectrum, without any dramatic turning points or sudden performance jumps. In comparison, IGNNK shows even less sensitivity to coverage, likely because it relies solely on traffic volume, while GNNUI and XGBoost benefit from being trained on diverse feature information. Nevertheless, GNNUI consistently outperforms IGNNK across all coverage levels.

Additionally, we find that not only the amount but also the selection of sensors included in the training data matters. Performance gains do not increase uniformly with coverage: for instance, in the Strava dataset, MAE increases slightly between 6\% and 9\% coverage before decreasing again. This variation suggests that GNNUI’s ability to leverage additional data depends on the informativeness and spatial distribution of the newly added segments. These results highlight the potential value of strategically placing traffic counters, an important direction for future work.



\section{Conclusion \label{sec:conclusion}}
Our experiments show that GNNUI consistently outperforms baselines, including IGNNK, a GNN-based interpolation model, and XGBoost, a machine learning method for feature-rich data. Our city-specific model adaptations, especially the integration of auxiliary node features, are key to the performance increases. We find that the choice of loss function also matters: the Zero-Inflated Negative Binomial loss achieves the best Kullback-Leibler divergence and competitive true-zero rates, crucial for distinguishing between high- and low-traffic streets for infrastructure planning. Meanwhile, GNLL and MAE losses yield lower MAE and RMSE, suggesting better average predictive performance. Depending on the data and application, different loss functions may be preferable for the task of interpolation.
Regarding graph structure, a binary adjacency matrix outperforms distance- and similarity-based alternatives across all metrics (MAE, RMSE, KL, true-zero rate). This likely reflects that binary adjacency matrices connect direct neighboring street segments, aligning with the strong local dependencies observed in real-world urban traffic patterns.
Finally, GNNUI achieves comparable performance to baselines while requiring considerably less training data. Across the full range of sensor availability, GNNUI shows only modest performance declines: for Strava, MAE increases from 7.1 to 10.5 when dropping from 90\% to 1\% coverage, and for Taxi, from 23.0 to 40.4. This highlights GNNUI’s strength in handling real-world urban conditions, where sensor infrastructure is typically very sparse, better than competing methods.

Our results suggest that sensor locations vary in informativeness, indicating that strategic sensor placement could enhance interpolation performance. Investigating this systematically remains an important avenue for future research. Further exploration into robust modeling techniques for urban areas with extremely limited sensor data is also warranted. We note that current benchmarking efforts for GNN-based traffic models are largely focused on highway and motorized traffic datasets. To better capture the complexity of urban environments, we advocate for systematically including diverse urban datasets and multiple transport modes in future benchmarking protocols. Lastly, future work could investigate augmenting the published Strava and taxi datasets to reduce inherent biases.

\newpage
\section*{Abbreviations}
\printglossary[type=\acronymtype, title={}, nonumberlist]

\newpage
\paragraph{Acknowledgements}
We are grateful the European Union’s Horizon Europe research and innovation program funded this project under Grant Agreement No 101057131, Climate Action To Advance HeaLthY Societies in Europe (CATALYSE). 

\paragraph{Author contributions}
Conceptualization \& Methodology: S.K.,  F.R., C.L.A., L.K.;
Formal Analysis \& Investigation: S.K.;
Software: S.K.;
Data Curation: S.K., 
Writing – Original Draft: S.K.,
Writing - Review \& Editing: S.K., C.L.A., F.R., L.K.;
Visualization: S.K.;
Funding acquisition: L.K.
All authors approved the final submitted draft.

\paragraph{Additional information}
Correspondence and requests for materials should be addressed to S.K. 

\paragraph{Competing Interests}
The authors declare no competing interests.

\paragraph{Availability of code}
The code and data supporting this study are made publicly available. The code is available here: \href{https://github.com/silkekaiser/GNNUI.git}{https://github.com/silkekaiser/GNNUI.git}. The data is available here: \cite{kaiser_data_2025}.

\appendix
\section*{Notation Used Throughout the Appendix}

This section summarizes the key variables used across the appendix. Most of these are not defined in the main text but are introduced in detail in Appendix~\ref{appx:trainingthroughmasking}. They are listed here for clarity and to avoid repetition.

\begin{itemize}
    \item $n_m$: number of masked nodes  
    \item $n_o$: number of observed nodes
    \item $T_s \in \mathbb{R}^{(n_o+n_m) \times h \times 1}$: target variable of a sample used during training 
    \item $\hat{T}_s \in \mathbb{R}^{(n_o+n_m) \times h \times 1}$: prediction of the target variable of a sample used during training 
    \item $T_s ^{M,N}\in \mathbb{R}^{(n_o+n_m) \times h \times 1}=T_s \circ M \circ N$:  target variable of a sample used during training with missing and masked nodes set to zero through multiplication with the respective matrices 
    \item $M_s \in \mathbb{R}^{(n_o+n_m)\times h \times 1}$: masking matrix of sample, equal to 0 if observation is masked 
    \item $M_t \in \mathbb{R}^{(n_s+n_u) \times h \times 1}$: masking matrix for test, equal to 0 if observation is masked 
        \item $N_s \in \mathbb{R}^{(n_o+n_m) \times h \times 1}$: missing matrix of sample, equal to 0 if observation is missing
    \item $N_t \in \mathbb{R}^{(n_s+n_u) \times h \times 1}$: missing matrix for test, equal to 0 if observation is missing 
\end{itemize}

\section{Training Loss and Error Computation \label{appx:training_loss}}
Throughout the appendix, we use several variables that are not defined in the main text but are necessary to explain the technical details across multiple appendix sections. To avoid repetition and ensure consistency, all such variables are defined once at the beginning of the appendix and referenced throughout, unless otherwise stated.

\textbf{Loss}
In GNNUI, we compute the loss on the masked locations only, while also accounting for missing observations. Where $\mathbf{1}$ is a matrix of ones of the same shape as mask matrix $M_{\text{s}}$: 

\begin{equation}
    \text{training loss masked only}= \text{loss}\left(\hat{T}_{\text{s}}, T_{\text{s}}; (\mathbf{1}- M_{\text{s}}) \circ N_{\text{s}} \right).
\end{equation}

In comparison, one could also compute the loss on masked and non-masked sensor, while accounting for missing observations (how IGNNK computes the loss):

\begin{equation}
    \text{training loss masked only}= \text{loss}\left(\hat{T}_{\text{s}}, T_{\text{s}}; N_{\text{s}} \right).
\end{equation}

The notation $\text{loss}(\hat{T}{\text{s}}, T{\text{s}}; (\mathbf{1}-M_{\text{s}}) \circ N_{\text{s}})$ indicates that the loss is computed element-wise, with each term weighted by the combined mask $(\mathbf{1}-M_{\text{s}}) \circ N_{\text{s}}$. This mask ensures that the loss is only calculated for relevant observations. It excludes non-masked data points (where $M_{\text{s}} = 1$, respectively $(\mathbf{1}-M_{\text{s}}) = 0$) and missing data points (where $N_{\text{s}} = 0$). 

For example, the MAE loss is:
\begin{equation}
    \text{MAE} = \frac{\sum (|\hat{T}_s - T_s| \circ ((\mathbf{1}-M_s) \circ N_s))}{\sum ((\mathbf{1}-M_s) \circ N_s)}
\end{equation}

This formulation ensures that only valid observations contribute to the loss calculation, preventing the loss from being artificially low due to non-masked or missing data.


\textbf{Error}
In the prediction step, we predict the traffic volumes for the unseen nodes. We denote the target feature in this testing phase as $T_t$, with the target feature of seen nodes denoted as $T_t^s$ and the target feature for unseen sensor denoted as $T^u_{t}$.

For the prediction of $T^u_{t}$, we create the corresponding matrix $W_t\in \mathbb{R}^{(n_t^s+n^u_t)\times (n_t^s+n^u_t)}$, the missing observations matrix $N_t\in \mathbb{R}^{(n_t^s+n^u_t) \times h \times 1}$ and the masking matrix $M_t\in \mathbb{R}^{(n_t^s+n^u_t) \times h \times 1}$. Here $M_t$ is equal to 0, if a node is unseen. Thus, $T_t =[T_t^s, T^u_{t}]$ and $X_t=[X_t^s, X^u_{t}]$.

We then feed $T_t^{M,N}=X_t\circ M_t \circ N_t$ and $X_t=[X_t^s, X^u_{t}]$ to the model to obtain $\hat{T}_t=[\hat{T_t^s}, \hat{T}^u_{t}] $ .

We split the test and validation data into subsets, each of size $h$, and compute each the prediction. We then compute the various errors across all predictions, only for unseen nodes, and taking missing observations into account.
\section{Training Through Masking Strategy \label{appx:trainingthroughmasking}}
Throughout the appendix, we use several variables that are not defined in the main text but are necessary to explain the technical details across multiple appendix sections. To avoid repetition and ensure consistency, all such variables are defined once at the beginning of the appendix and referenced throughout, unless otherwise stated.

To train the model, we simulate the interpolation process by constructing multiple training samples. A sample is a small training instance that includes a) a selected time window of length $h$, and b) a subset of nodes, randomly divided into supposedly seen (\textit{observed}, i.e., equipped with sensors) and supposedly unseen (\textit{masked nodes}, i.e., to be interpolated). Each sample mimics a localized spatio-temporal snapshot of the network, allowing the model to repeatedly learn the interpolation task from different perspectives.
During training, we iterate over $S = p / (h \times \text{batch size})$ samples in each epoch, where $p$ is the total number of time steps. Each sample corresponds to a unique time window beginning at index $t \in [1, p-h]$, ensuring that time windows do not overlap. This differs from \cite{wu_inductive_2021}, where time points are randomly sampled and may overlap. By ensuring that time windows do not overlap, we guarantee that the full training dataset is covered within each epoch. 

Within each sample, we randomly choose $n_o$ street segments to be treated as observed, and $n_m$ as masked, such that $n_o + n_m \leq n$. Allowing each sample to cover only part of the graph improves generalization and reduces computational cost. We construct the training sample tensors ($X_s$ and $T_s$) as well as the related adjacency ($W_s$) by slicing the full node feature and target tensors according to the selected nodes and time window:
\begin{align}
    &X_{\text{s}} = X[\{i^1, \cdots, i^{n_0}, \cdots, i^{n_0+n_m}\}, [t, t+h),:] \nonumber \\
    &T_{\text{s}} = T[\{i^1, \cdots, i^{n_0}, \cdots, i^{n_0+n_m}\}, [t, t+h)] \nonumber \\
    &W_{\text{s}} \in \mathbb{R}^{(n_o+n_m) \times (n_o+n_m)}.
\end{align} 

As in the test setting, features of observed and masked nodes are known, the target variable is only known for the observed nodes. To simulate that the target variables of the masked nodes are missing, we use the mask matrix $M_s \in \mathbb{R}^{(n_o+n_m)\times h \times 1}$, where the value 1 indicates that the node is observed and 0 otherwise. 
 Let's call the set of $n_m$ selected masked nodes $\mathcal{I}_m$. Then the entries of $M_{\text{s}}$ are defined as:
\begin{equation}
M_{\text{s}}[i,:] =
\begin{cases}
0, & \text{if } i \in \mathcal{I}_m, \\
1, & \text{otherwise.}
\end{cases}
\end{equation}

During training, we multiply the target variables $T_{\text{s}}$ with the mask matrix $M_{\text{s}}$ element-wise, which sets the target value for the masked observations to zero, effectively hiding the target variable of masked nodes. 

We also use masking for missing values. Missing values are common in real-world sensor data and can be caused by malfunctions, maintenance, or construction activities. For instance, after removing outliers and accounting for gaps, 1.3\% of the Strava data and 1.0\% of the taxi data are missing. We handle the missing data by constructing another mask matrix, $N_{\text{s}}\in \mathbb{R}^{(n_o+n_m)\times h \times 1}$:
The entries of $N_{\text{s}}$ are defined as:
\begin{equation}
N_{\text{s}}[i,\Tilde{t}]=
    \begin{cases}
      0, & \text{if}\ \text{the measurement at }X_{\text{s}}[i,\Tilde{t}]\text{ is missing,} \\
      1, & \text{otherwise.}
    \end{cases}
\end{equation}

We element-wise multiply the target variable $T_{\text{s}}$ with $N_{\text{s}}$, effectively setting missing target variables to zero. The target variable of the sample is $T^{M,N}_{\text{s}} = T_{\text{s}} \circ M_{\text{s}}\circ N_{\text{s}}$, with $\circ$ denoting element-wise multiplication.

To enable the model to distinguish between true zero values, masked values, and missing values, all represented numerically as zero, we explicitly provide this information as node features to the model. Specifically, the two binary matrices are included as two channels in the node feature tensor \( X_s \). Let \( X_s^{\text{raw}} \in \mathbb{R}^{(n_o + n_m) \times h \times k^{\text{raw}}} \) denote the feature tensor without the two binary indicators. We augment it by appending \( M_s \) (indicating masked values) and \( N_s \) (indicating missing values) as additional channels along the feature dimension. The resulting input tensor becomes:
\[
X_s = \text{concat}(X_s^{\text{raw}}, M_s, N_s) \in \mathbb{R}^{(n_o + n_m) \times h \times k}
\tag{B.4}
\]
where \( k = k^{\text{raw}} + 2 \) reflects the inclusion of the two binary indicators. We include these binary features in the exact same manner during inference.

Thus, for each sample we train the model to predict $\hat{T}_\text{s} \in \mathbb{R}^{(n_o+n_m)\times h \times 1}$ given $X_{\text{s}} \in \mathbb{R}^{(n_o+n_m)\times h \times k}$ and $T^{M,N}_{\text{s}}\in \mathbb{R}^{(n_o+n_m)\times h \times 1}$.

\section{Temporal Generalization Ablation Study \label{appx:temporal_generalization}}
\begin{figure}[hb!]
    \centering    
    \includegraphics[width=0.6\textwidth]{02_graphs/GNNUI_extrapolating_legend.png}
    \vspace{0.5cm} 

    \begin{subfigure}[b]{0.45\textwidth}
            \centering
        \includegraphics[width=\textwidth]{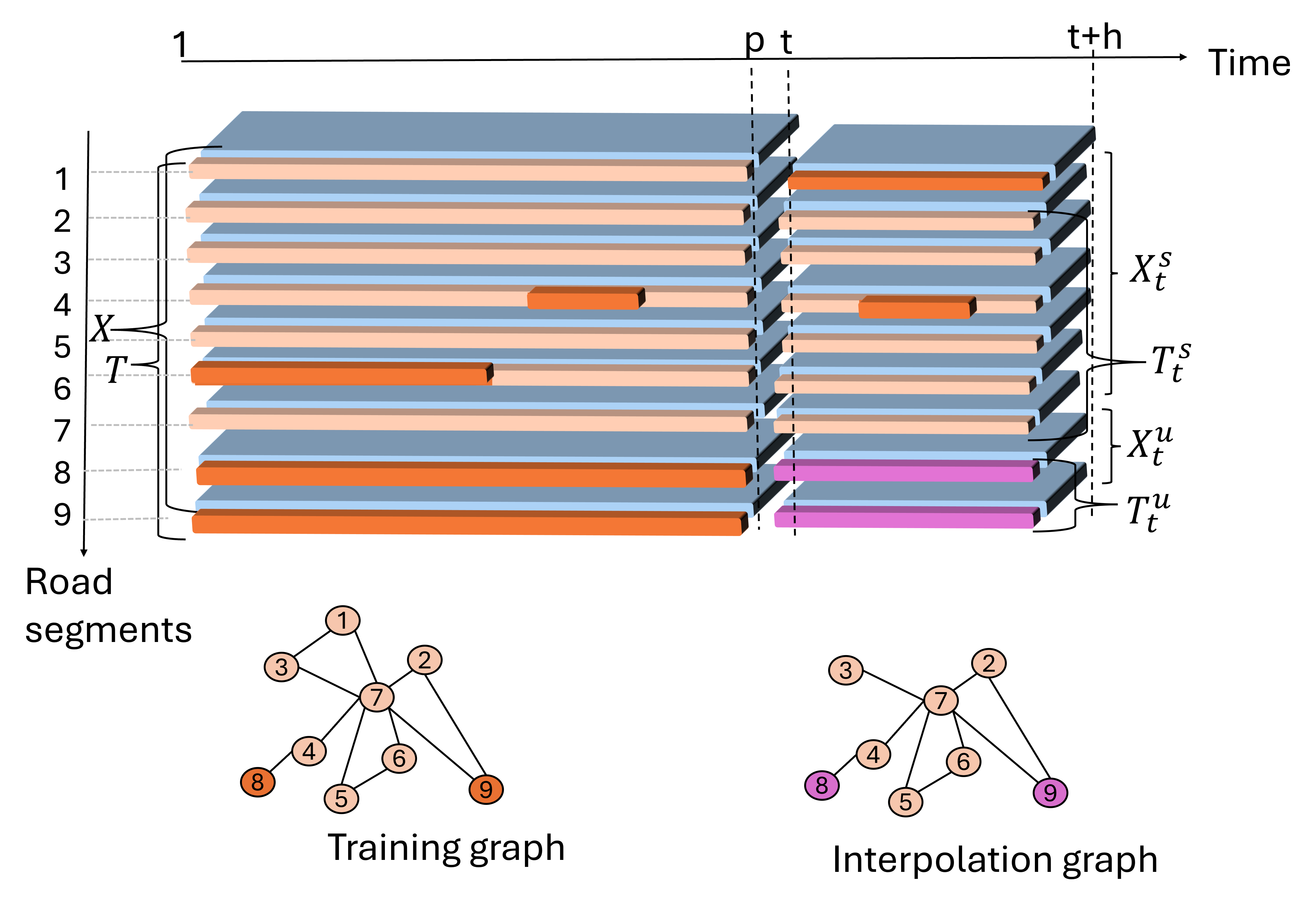}
        \caption{Using the entire graph in training.}
        \label{fig:explanaion_interpolation_b}
    \end{subfigure}
    \hfill
    \begin{subfigure}[b]{0.45\textwidth}
        \centering
        \includegraphics[width=\textwidth]{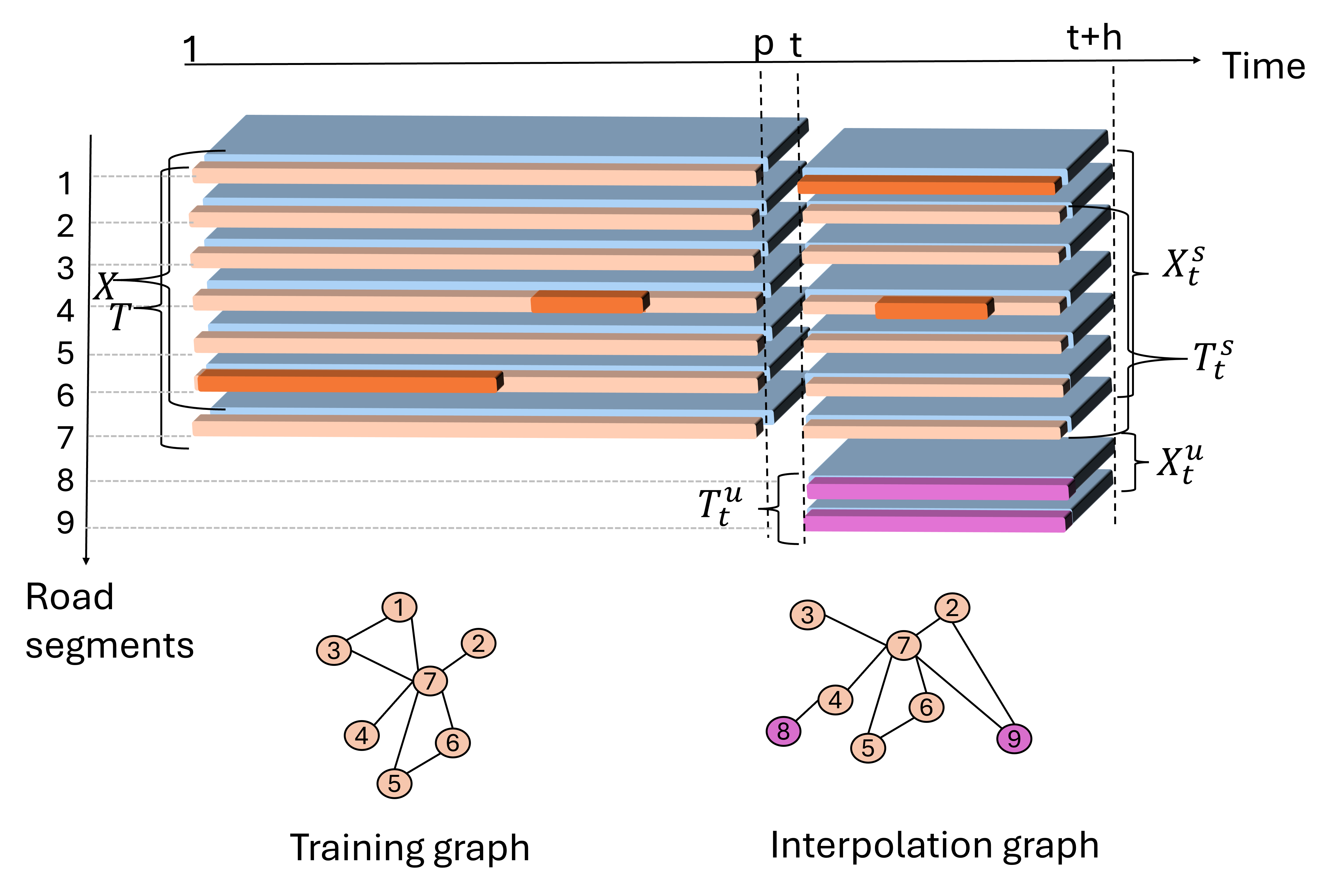}
        \caption{Without using the entire graph in training.}
        \label{fig:explanaion_interpolation_d}
    \end{subfigure}
    \caption{The figure corresponds to Figure \ref{fig:explanaion_interpolation} in the main text - but here in a setup, where the prediction step lies in the future of the training.}
    \label{fig:explanaion_interpolation_appendix}
\end{figure}

Throughout the appendix, we use several variables that are not defined in the main text but are necessary to explain the technical details across multiple appendix sections. To avoid repetition and ensure consistency, all such variables are defined once at the beginning of the appendix and referenced throughout, unless otherwise stated.

In the main paper, we design our test and training data such that the test period overlaps with the training period, meaning the model has partial access to test-time inputs. In contrast, a second setup requires the model to generalize to both unseen locations and future times, where no test inputs are observed during training.

For example, the model is trained on street segments 1 to 9 during the time interval $1$ to $p$ (target value of 8 and 9 set to missing) and tested on street segments 8 and 9 in a future time ($[t,t+h)$, where $t > p$). Unlike in the main paper, no input for the test set (e.g., $X^s_t$, $T^{s,N}_t=T_t^s\circ N_s$, nor $X_t^{u}$) was seen during. This task requires spatial interpolation as well as forecasting, and presents a greater challenge than only spatial interpolation. This is valuable in applications requiring continuously updated traffic forecasts, such as dynamic traffic control. We depict this setup in Figure \ref{fig:explanaion_interpolation_appendix}, for both the case when the entire graph is utilized in training and when it is not.

In comparison to the setup in the main text, we do not only need to do a spatial split when creating the training/validation/test set, but also a temporal split. Here, we split the dataset chronologically: the first 70\% of time steps are used for training, the next 15\% for validation, and the final 15\% for testing. 

For this set-up, we also provide the results of the ablation study (\ref{Table:table1_interpolation_appx_appx}). Since the setup in the main text uses a different train/validation/test split, direct comparisons between the observed errors are not meaningful.

\begin{table}[h]
\centering
    \caption{This Table corresponds to Table \ref{Table:table1_interpolation_appx} in the main text - but here in a setup, where the prediction step lies in the future of the training. N/A indicates that the model does not produce a probabilistic output required to compute NLL.}
    \label{Table:table1_interpolation_appx_appx}
    \tiny
        \begin{tabular}{lccccc|rrrrr|rrrrr}
            \toprule
            \multicolumn{1}{l}{}&  \multicolumn{5}{c}{Components}& \multicolumn{5}{c}{Strava}&   \multicolumn{5}{c}{Taxi}\\
            \midrule
            & \rotatebox{90}{Entire graph (i) }& \rotatebox{90}{Node features (ii)} & \rotatebox{90}{Loss on masked only (iii)}   &\rotatebox{90}{Feature missing/masked (iv)} &\rotatebox{90}{training loss (v)}&     &  &  &  & &  &  &  &    \\
            &     &  &   &   &    &MAE  &RMSE  & KL & Zero &NLL &MAE  &RMSE  & KL & Zero &NLL \\
            \midrule
IGNNK &   &   &   &   & MSE & 12.40 & 21.4 & 0.22 & 0.058 & N/A & 35.58 & 73.3 & 0.22 & 0.189 & N/A \\
XGBoost &   &   &   &   & MAE & 9.77 & 19.1 & 0.36 & 0.580 & N/A & 26.94 & 74.1 & 0.36 & 0.668 & N/A \\
            \cline{2-16}
            GNNUI & \checkmark & \checkmark & \checkmark & \checkmark & ZINB & \textbf{8.39} & 16.3 & 0.16 & \textbf{0.661} & 0.650\textsuperscript{*} & 25.55 & 58.4 & \textbf{0.03} & 0.744 & 0.635\textsuperscript{*} \\
  & \checkmark & \checkmark & \checkmark & \checkmark & NB & 8.72 & \textbf{16.2} & \textbf{0.13} & 0.502 & 0.266\textsuperscript{*} & 25.65 & 57.3 & 0.05 & 0.591 & 0.333\textsuperscript{*} \\
              \cline{2-16}

  & \checkmark & \checkmark & \checkmark & \checkmark & GNLL & 9.04 & 17.0 & 0.69 & 0.265 & 278 & 20.76 & 49.0 & 0.22 & 0.296 & 2390 \\
  & \checkmark & \checkmark & \checkmark &   & GNLL & 9.93 & 19.9 & 2.23 & 0.282 & 382 & 21.20 & \textbf{47.0} & 0.19 & 0.221 & 2200 \\
  & \checkmark & \checkmark &   &   & GNLL & 9.95 & 20.6 & 1.66 & 0.268 & 413 & 23.03 & 53.3 & 0.20 & 0.359 & 2837 \\
  & \checkmark &   &   &   & GNLL & 11.71 & 21.3 & 1.85 & 0.056 & 433 & 38.12 & 75.4 & 0.31 & 0.004 & 5624 \\
  &   &   &   &   & GNLL & 11.41 & 20.9 & 1.66 & 0.042 & 418 & 37.31 & 73.9 & 0.49 & 0.009 & 5394 \\
            \cline{2-16}
  & \checkmark & \checkmark & \checkmark & \checkmark & MAE & 8.42 & 17.0 & 0.48 & 0.595 & N/A & 21.74 & 60.2 & 0.36 & 0.396 & N/A \\
  & \checkmark & \checkmark & \checkmark &   & MAE & 9.76 & 19.3 & 1.18 & 0.367 & N/A & \textbf{20.38} & 53.7 & 0.21 & \textbf{0.748} & N/A \\
  & \checkmark & \checkmark &   &   & MAE & 10.51 & 20.7 & 1.59 & 0.263 & N/A & 30.38 & 90.1 & 2.12 & 0.219 & N/A \\
  & \checkmark &   &   &   & MAE & 12.24 & 24.8 & 3.22 & 0.315 & N/A & 35.15 & 99.1 & 3.38 & 0.638 & N/A \\
  &   &   &   &   & MAE & 12.27 & 24.9 & 3.22 & 0.368 & N/A & 35.29 & 99.2 & 3.84 & 0.405 & N/A \\
  \bottomrule
    \end{tabular}
        \vspace{1ex}
    \begin{minipage}{\textwidth}
    \tiny
    \textsuperscript{*}NLL values from GNLL, ZINB, and NB are not directly comparable, as they are computed over different types of probability distributions.
    \end{minipage}
    
\end{table}

\section{Model Pipeline \label{appx:model_pipeline}}
Throughout the appendix, we use several variables that are not defined in the main text but are necessary to explain the technical details across multiple appendix sections. To avoid repetition and ensure consistency, all such variables are defined once at the beginning of the appendix and referenced throughout, unless otherwise stated.

The neural network consists of three Diffusion Graph Convolutional Network (DGCN) layers.
The input to the first layer consists of a combination of the feature nodes and the target variables. Specifically, we construct $H_0$ by first concatenating the sampled feature matrix $X_{\text{s}}$ with the target tensor $T_{\text{s}}^{M,N}$ along the third axis. This combined matrix is then reshaped into a two-dimensional structure, where the first dimension represents the number of nodes $(n_o + n_m)$, and the second dimension corresponds to the number of features plus one, times the time dimension $(h(k+1))$.

The output of the $l^{th}$ DGCN layer $H_{l+1}$ is:
\begin{equation}
    H_{l+1} = \sum_{k=1}^K T_k(\Bar{W}_f)H_l\Theta_{f,l}^k+T_k(\Bar{W}_b)H_l\Theta_{b,l}^k, 
\end{equation}
 $\Bar{W}_f = W_{s}/\text{rowsum}(W_{s})$ and $\Bar{W}_b = W_{s}^T/\text{rowsum}(W_{s})^T$ are the forward and backward transition matrix. This allows the incorporation of directed graphs in this model. With K as the order of convolution, 
 the Chebyshev polynomial $T_k(X)$ is used to approximate the graph convolutional process. $\Theta_{b,l}^k$ and $\Theta_{f,l}^k$ are learning parameters. 
 
The second layer output $H_2$ includes a residual connection from $H_1$. This allows information from the previous layer to be retained and carried forward. We set the hidden feature dimension to $z$. For the first and second layers, we employ a ReLU activation function. For the third layer, we use a linear activation function to obtain $H_3$. We also apply layer normalization for all three layers to stabilize training. 
 
\begin{figure}[!ht]
    \centering
    \includegraphics[width=\textwidth]{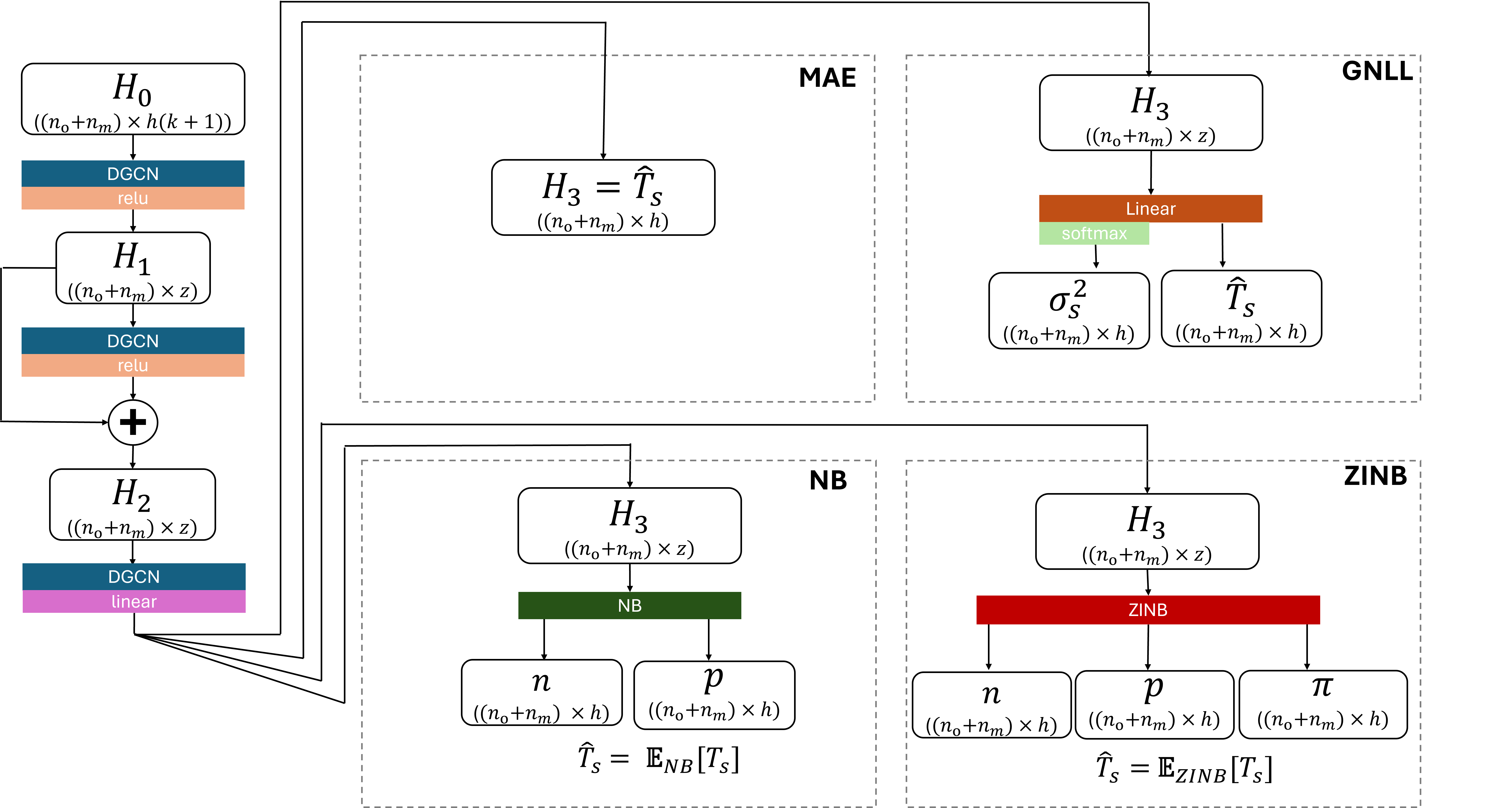}
    \caption{The pipeline architecture of our models varies depending on the chosen loss function (\acrshort{mae}, \acrshort{gnll}, \acrshort{nb}, or \acrshort{zinb}). All variants share a common structure up to layer $H_3$, shown on the left side of the figure. The right side illustrates the diverging final layers. Each layer outputs those parameters needed for the loss computation. As for the prediction itself, for \acrshort{mae} and \acrshort{gnll}, the model directly predicts $\hat{T}_s$. For \acrshort{nb} and \acrshort{zinb}, $\hat{T}_s$ is derived from the expected value of the respective probabilistic distribution.}
    \label{fig:pipeline}
\end{figure}

The rest of the model's design depends on what type of loss is used. To implement the different losses, we need to obtain different parameters. This is done through the last layer. To compute the \acrshort{mae} loss, we only need a prediction and $\hat{T}_{\text{s}}=H_3$. In the case of \acrshort{gnll} we need both a variance and mean. Thus, we employ a linear layer as the output layer to $H_3$ to obtain the mean, which is the prediction $\hat{T}_{\text{s}}$ and a linear layer with an softmax activation function to obtain the variance $\sigma^2_{\text{s}}$.
For \acrshort{nb} we pass $H_3$ through an NB layer as the output layer to obtain $n$ and $p$. For and \acrshort{nb} we pass $H_3$ through an NB layer to obtain $n$, $p$ and $\pi$. The NB layer applies two separate convolutional operations to parameterize the distribution: 
\begin{align} n &= \text{softplus}(\text{Conv}(H_3, \theta_{out,n}) + \beta_{out,n}),\\
\ p &= \text{sigmoid}(\text{Conv}(H_3, \theta_{out,p}) + \beta_{out,p}).
\end{align} 
The ZINB layer additionally applies:
\begin{equation}
    \pi = \text{sigmoid}(\text{Conv}(H_3, \theta_{out,\pi}) + \beta_{out,\pi}).
\end{equation}
where $\text{Conv}(\cdot,\theta_{out})$ is a 1×1 convolution, $\theta_{out}$ are the learned weights, and $\beta_{out}$ are the biases. The activations ensure that:  $n>0$, $p\in(0,1)$,  and  $\pi \in (0,1)$. To be able to compute error metrics, that require a prediction, we can obtain the prediction $\hat{T}_{\text{s}}$ for both \acrshort{nb} and \acrshort{zinb} as the expected value of their respective distributions using these parameters: 
   $ \hat{T}_{\text{s}}=\mathbb{E}_{NB}[T_{\text{s}}]=n(1-p)/p$ and  $\hat{T}_{\text{s}}=\mathbb{E}_{ZINB}[T_{\text{s}}]=(1-\pi)n(1-p)/p$. 
\section{Zero Inflated Negative Binomial and Negative Binomial Loss \label{appx:zinb_and_nb}}
Throughout the appendix, we use several variables that are not defined in the main text but are necessary to explain the technical details across multiple appendix sections. To avoid repetition and ensure consistency, all such variables are defined once at the beginning of the appendix and referenced throughout, unless otherwise stated.

The Gaussian Negative Log-Likelihood, for a predicted mean $\hat{T}_{\text{s}}$ and variance $\sigma^2_{\text{s}}$, is given by: 
 
 \begin{align}
NLL_{\text{Gaussian}} = \frac{1}{2} \left( \frac{(T_{\text{s}} - \hat{T}_{\text{s}})^2}{\max(\sigma_{\text{s}}^2, \epsilon)} + \log(\max ( \sigma_{\text{s}}^2, \epsilon)) \right) \circ (\mathbf{1}-M_{\text{s}}) \circ N_{\text{s}}.
\end{align}
where $\epsilon=1\times 10^{-6}$ ensures numerical stability. 

\textbf{NB loss}.
The probability mass function of a variable $T$ following the \acrshort{nb} distribution is given by:

\begin{equation}
    f_{NB}(T; n, p)  = \binom{T + n - 1}{n - 1} (1 - p)^{T} p^n.
\end{equation}
Where $n$ and $p$ are shape parameters representing the number of successes and the probability of a single failure, respectively.

\textbf{ZINB loss}.
To account for the high number of zeros in the data, the \acrshort{zinb} distribution introduces an additional parameter, $\pi$, to explicitly model the inflation of zeros. The probability mass function for the \acrshort{zinb} distribution is:

\begin{equation}
f_{ZINB}(T; \pi, n, p) = 
\begin{cases} 
\pi + (1 - \pi) f_{NB}(0; n, p) & \text{if } T = 0, \\
(1 - \pi) f_{NB}(T; n, p) & \text{if } T > 0.
\end{cases}    
\end{equation}
Here the data distribution is modeled as zeros with a probability of $\pi$, or as non-zero values following the \acrshort{nb} distribution with a probability of $(1-\pi)$.  $\pi$, $n$ and $p$ are parameterized by the GNN.

The loss is then calculated as the negative log likelihood. For \acrshort{nb}:
\begin{align}
 NLL_{NB}=(\Gamma(n)+\Gamma(T_{\text{s}}+1)-\Gamma(n+T_{\text{s}}) -n\log(p)-T_{\text{s}}\log(1-p)) \circ (\mathbf{1}-M_{\text{s}}) \circ N_{\text{s}}.
\end{align}
respectively, if computing the loss both on masked and non-masked observations:
\begin{align}
 NLL_{NB}=(\Gamma(n)+\Gamma(T_{\text{s}}+1)-\Gamma(n+T_{\text{s}}) -n\log(p)-T_{\text{s}}\log(1-p))  \circ N_{\text{s}}.
\end{align}
and the log likelihood for \acrshort{zinb}:
\begin{equation}
LL_{T_{\text{s}}} =
\begin{cases} 
(\log \pi + \log (1 - \pi)p^n) \circ (\mathbf{1}-M_{\text{s}}) \circ N_{\text{s}}, & \text{when } T_{\text{s}} = 0, \\
(\log (1 - \pi) + \log \Gamma(n + T_{\text{s}}) - \log \Gamma(T_{\text{s}} + 1) \\
\quad - \log \Gamma(n) + n \log p + T_{\text{s}} \log (1 - p))\circ (\mathbf{1}-M_{\text{s}}) \circ N_{\text{s}}, & \text{when } T_{\text{s}} > 0.
\end{cases}
\end{equation}

respectively,

\begin{equation}
LL_{T_{\text{s}}} =
\begin{cases} 
(\log \pi + \log (1 - \pi)p^n)\circ  N_{\text{s}}, & \text{when } T_{\text{s}} = 0, \\
(\log (1 - \pi) + \log \Gamma(n + T_{\text{s}}) - \log \Gamma(T_{\text{s}} + 1)\\
\quad - \log \Gamma(n) + n \log p + T_{\text{s}} \log (1 - p)) \circ N_{\text{s}}, & \text{when } T_{\text{s}} > 0.
\end{cases}
\end{equation}
With the negative log likelihood:
\begin{equation}
NLL_{ZINB} =  -LL_{T_{\text{s}}=0} -LL_{T_{\text{s}}>0}.
\end{equation}

\section{Errors}\label{apx:errors}
Throughout the appendix, we use several variables that are not defined in the main text but are necessary to explain the technical details across multiple appendix sections. To avoid repetition and ensure consistency, all such variables are defined once at the beginning of the appendix and referenced throughout, unless otherwise stated.


\begin{equation}
    \text{MAE} = \frac{\sum (|\hat{T}_t - T_t| \circ ((\mathbf{1}-M_t) \circ N_t))}{\sum ((\mathbf{1}-M_t) \circ N_t)}
\end{equation}


\begin{equation}
    RMSE = \sqrt{\frac{\sum ((T_{\text{t}} - \hat{T}_{\text{t}})^2\circ ((\mathbf{1}-M_t) \circ N_t))}{\sum ((\mathbf{1}-M_t) \circ N_t)}}
\end{equation}





\section{True Zero Rate \label{appx:true_zero_rate}}
 Throughout the appendix, we use several variables that are not defined in the main text but are necessary to explain the technical details across multiple appendix sections. To avoid repetition and ensure consistency, all such variables are defined once at the beginning of the appendix and referenced throughout, unless otherwise stated.
 
\begin{equation}
\text{true-zero rate} = \frac{\sum  ( \mathbbm{1}(T_{\text{t}} = 0) \circ \mathbbm{1}(\mid \hat{T}_{\text{t}} \mid < \tau)\circ(1-M_t)\circ N_t)}{\sum    (\mathbbm{1}(T_{\text{t}} = 0)\circ(1-M_t)\circ N_t)}
\end{equation}
$\mathbbm{1}$ is an indicator function equal to 1 if true. We set $\tau=0.99$, to allow for small deviations around zero. Our aim is not only to identify exact zeros, but also to capture very low values characteristic of sparse urban traffic patterns. These near-zero values are often as important as exact zeros in representing inactivity or minimal flow. By using a slightly higher threshold, we ensure the model is evaluated on its ability to approximate such low-traffic conditions.
\clearpage

\section{Strava Descriptive Statistics \label{apx:strava_descriptives}}

\begin{table}[h]
    \centering
    \begin{tabular}{l|r}
    \toprule
            Type & Percentage share\\
            \midrule
            \multicolumn{2}{l}{Gender}\\
            \midrule
              Female	&11.44  \% \\
              Male	&87.55  \% \\
              Unspecified	&1.01  \% \\
            \midrule
            \multicolumn{2}{l}{Time of day}\\
            \midrule
              05:00 - 10:00 &22.89  \% \\
              10:00 - 15:00	&29.95  \% \\
              15:00 - 20:00	&40.666  \% \\
              20:00 - 05:00	&6.50  \% \\
            \midrule
            \multicolumn{2}{l}{Age}\\
            \midrule
              18-34	&36.14  \% \\
              35-54	&59.52  \% \\
              55-64	&4.08  \% \\
              65+	&0.26  \% \\
            \midrule
            \multicolumn{2}{l}{Purpose}\\
            \midrule
              Commute	&38.59  \% \\
              Leisure	&61.41  \% \\
            \midrule
            \multicolumn{2}{l}{Type of bike}\\
            \midrule
              Non-e-bike	&99.75  \% \\
              E-bike	&0.25 \%\\
            \midrule
            \multicolumn{2}{l}{Average speed}\\
            \midrule
            Average speed	&26.27 km/h \\
            \bottomrule
    \end{tabular}
    \caption{Depicted are the shares of various characteristics among the Strava data used (Berlin, 2019-2023, based on  daily observations for the cycling network). It becomes apparent that the data is mainly generated by male, young, and middle-aged users cycling fast on non-e-bikes.}
    \label{tab:my_label}
\end{table}
\section{Node Features}\label{apx:list_of_node_features}

\begin{landscape}
\begin{tiny}
\begin{longtable}{|p{4cm}|p{1cm}|p{1.5cm}|p{1cm}|p{1cm}|p{1cm}|p{2cm}|p{2cm}|p{2cm}|p{1cm}|p{1cm}|}
    \caption{Node Features} \label{tab:node_features} \\
    \toprule
    Feature & Further information & Spatial scope & \multicolumn{2}{c|}{Temporal scope} & \# features & Type & \multicolumn{2}{c|}{References} & \multicolumn{2}{c|}{Included in} \\
    \cline{ 4-5}\cline{8-11}
    & & &Strava & Taxi& & & Strava & Taxi & Strava & Taxi \\

 \hline
\multicolumn{11}{|l|}{Infrastructure indicators}\\
 \hline
Latitude and longitude & &street-level &\multicolumn{2}{|l|}{constant}&2&numerical&\cite{senate_department_for_the_environment_mobility_consumer_and_climate_protection_berlin_jahresdatei_2023, openstreetmap_contributors_planet_2017}& \citep{boeing_modeling_2025}&\checkmark& \checkmark\\
Number of education facilities&   &within a radius of 50m, 100m, 250m and 500m &\multicolumn{2}{|l|}{constant}&4&numerical&\multicolumn{2}{|l|}{\cite{openstreetmap_contributors_planet_2017}}&\checkmark& \checkmark \\
Number of hospitals&   &within a radius of 50m, 100m, 250m, and 500m &\multicolumn{2}{|l|}{constant}&4&numerical&\multicolumn{2}{|l|}{\cite{openstreetmap_contributors_planet_2017}} &\checkmark& \checkmark \\
Number of shops&   &within a radius of 50m, 100m, 250m, and 500m &\multicolumn{2}{|l|}{constant}&4&numerical&\multicolumn{2}{|l|}{\cite{openstreetmap_contributors_planet_2017}}&\checkmark& \checkmark \\
Number of industrial facilities&   &within a radius of 50m, 100m, 250m, and 500m &\multicolumn{2}{|l|}{constant}&4&numerical&\multicolumn{2}{|l|}{\cite{openstreetmap_contributors_planet_2017}} &\checkmark& \checkmark \\
Number of hotels &   &within a radius of 50m, 100m, 250m, and 500m &\multicolumn{2}{|l|}{constant}&4&numerical&\multicolumn{2}{|l|}{\cite{openstreetmap_contributors_planet_2017}} &\checkmark& \checkmark \\
Number of restaurants &    &within a radius of 50m, 100m, 250m and 500m &&constant&4&numerical&&\cite{openstreetmap_contributors_planet_2017} && \checkmark \\
Number of bars &  &within a radius of 50m, 100m, 250m and 500m &&constant&4&numerical&&\cite{openstreetmap_contributors_planet_2017} && \checkmark \\
Number of cafes &   &within a radius of 50m, 100m, 250m and 500m &&constant&4&numerical&&\cite{openstreetmap_contributors_planet_2017} && \checkmark \\
Number of railway stations &    &within a radius of 50m, 100m, 250m and 500m &&constant&4&numerical&&\cite{openstreetmap_contributors_planet_2017} && \checkmark \\
Number of bus stops &   &within a radius of 50m, 100m, 250m and 500m &&constant&4&numerical&&\cite{openstreetmap_contributors_planet_2017} && \checkmark \\
Number of long-distance bus stops &   &within a radius of 50m, 100m, 250m and 500m &&constant&4&numerical&&\cite{openstreetmap_contributors_planet_2017} && \checkmark \\
Distance to the city center  &in km & street-level &\multicolumn{2}{|l|}{constant}&1&numerical&\multicolumn{2}{|l|}{\cite{openstreetmap_contributors_planet_2017}} &\checkmark& \checkmark \\
Type of bicycle lane & $\in$ \{no bicycle lane, bicycle lane on sidewalk, bicycle lane on street\} & street-level &constant & &1&categorical&\cite{openstreetmap_contributors_planet_2017}& &\checkmark&  \\
Type of street  & $\in$ \{primary, residential, secondary, pedestrian, path, tertiary, unknown\} & street-level &constant&&1&categorical&\cite{openstreetmap_contributors_planet_2017} &&\checkmark&  \\
Type of street  & $\in$ \{secondary, residential, primary, tertiary, motorway link, trunk,  motorway, trunk link, tertiary link, primary link, secondary link, living street, unknown\} & street-level &&constant&1&categorical&&\cite{openstreetmap_contributors_planet_2017} && \checkmark \\
Street surface & $\in$ \{ unknown, asphalt, sett, concrete, paving-stones, paved\} 
& street-level &\multicolumn{2}{|l|}{constant}&1&categorical&\multicolumn{2}{|l|}{\cite{openstreetmap_contributors_planet_2017}} &\checkmark& \checkmark \\
Number of car lanes & $\in$ \{unknown, 1, 2, $\cdots$\}&  street-level &&constant&1&categorical&&\cite{openstreetmap_contributors_planet_2017} &\checkmark& \checkmark \\
One way street & & street-level &&constant&1&binary&&\cite{openstreetmap_contributors_planet_2017} && \checkmark \\
Max. permitted speed for cars  & in km/h,  $\in$ \{0, 30, 50\}&street-level &constant&&1&categorical&\cite{openstreetmap_contributors_planet_2017}& &\checkmark&  \\
Max. permitted speed for cars  & in mph,  $\in$ \{unknown, 20 mph or less, 15 mph, 20 mph, 25 mph, 30 mph, 35 mph, 40 mph, 45 mph, 50 mph\}&street-level &&constant&1&categorical&&\cite{openstreetmap_contributors_planet_2017} && \checkmark \\
Parking  &   $\in$ \{yes, no, unknown\}&street-level &&constant&1&categorical&&\cite{openstreetmap_contributors_planet_2017} && \checkmark \\
Cyclability  &  &street-level &&constant&1&numerical&\cite{openstreetmap_contributors_planet_2017}& & \checkmark& \\
Cyclability (commute) &  &street-level &&constant&1&numerical&\cite{openstreetmap_contributors_planet_2017}& &\checkmark&  \\
Cyclability (touring) &  &street-level &&constant&1&numerical&\cite{openstreetmap_contributors_planet_2017} &&\checkmark&  \\
Part of long distance cycling route &  &street-level &&constant&1&numerical&\cite{openstreetmap_contributors_planet_2017} &&\checkmark&  \\
Arable land&\% of planning area used for &planning area&constant&&1&numerical&\cite{senate_department_for_urban_development_building_and_housing_lebensweltlich_2023}& &\checkmark&  \\
Horticulture& \% of planning area used for&planning area&constant&&1&numerical&\cite{senate_department_for_urban_development_building_and_housing_lebensweltlich_2023}& &\checkmark&  \\
Cemetery&\% of planning area used for &planning area&constant&&1&numerical&\cite{senate_department_for_urban_development_building_and_housing_lebensweltlich_2023}& &\checkmark&  \\
Public use& \% of planning area used for&planning area&constant&&1&numerical&\cite{senate_department_for_urban_development_building_and_housing_lebensweltlich_2023}& &\checkmark&  \\
Waterways& \% of planning area used for&planning area&constant&&1&numerical&\cite{senate_department_for_urban_development_building_and_housing_lebensweltlich_2023}& &\checkmark&  \\
Business& \% of planning area used for&planning area&constant&&1&numerical&\cite{senate_department_for_urban_development_building_and_housing_lebensweltlich_2023}& &\checkmark&  \\
Pasture& \% of planning area used for&planning area&constant&&1&numerical&\cite{senate_department_for_urban_development_building_and_housing_lebensweltlich_2023}& &\checkmark&  \\
Allotment garden&\% of planning area used for &planning area&constant&&1&numerical&\cite{senate_department_for_urban_development_building_and_housing_lebensweltlich_2023}& &\checkmark&  \\
Park& \% of planning area used for&planning area&constant&&1&numerical&\cite{senate_department_for_urban_development_building_and_housing_lebensweltlich_2023}& &\checkmark&  \\
Town squares&\% of planning area used for &planning area&constant&&1&numerical&\cite{senate_department_for_urban_development_building_and_housing_lebensweltlich_2023}& &\checkmark&  \\
Waste disposal& \% of planning area used for&planning area&constant&&1&numerical&\cite{senate_department_for_urban_development_building_and_housing_lebensweltlich_2023}& &\checkmark&  \\
Traffic area&\% of planning area used for &planning area&constant&&1&numerical&\cite{senate_department_for_urban_development_building_and_housing_lebensweltlich_2023}& &\checkmark&  \\
Forest&\% of planning area used for &planning area&constant&&1&numerical&\cite{senate_department_for_urban_development_building_and_housing_lebensweltlich_2023}& &\checkmark&  \\
Weekend housing&\% of planning area used for &planning area&constant&&1&numerical&\cite{senate_department_for_urban_development_building_and_housing_lebensweltlich_2023}& &\checkmark&  \\
Residential housing& \% of planning area used for&planning area&constant&&1&numerical&\cite{senate_department_for_urban_development_building_and_housing_lebensweltlich_2023}& &\checkmark&  \\
 \hline
 \multicolumn{11}{|l|}{Weather indicators}\\
 \hline
Average  temperature &in C°&city-wide&daily&daily \& hourly &1       $\mid$ 2&numerical&\multicolumn{2}{|l|}{\cite{meteostat_weathers_2022}} &\checkmark& \checkmark \\
Minimium daily temperature &in C°&city-wide&daily&daily&1&numerical&\multicolumn{2}{|l|}{\cite{meteostat_weathers_2022}} &\checkmark& \checkmark  \\
Maximum daily temperature &in C°&city-wide&daily&daily&1&numerical&\multicolumn{2}{|l|}{\cite{meteostat_weathers_2022}} &\checkmark& \checkmark  \\
Precipitation &in mm &city-wide&daily&daily \& hourly &1   $\mid$ 2&numerical&\multicolumn{2}{|l|}{\cite{meteostat_weathers_2022}} &\checkmark& \checkmark  \\
Maximum snow depth & in mm& city-wide&daily&daily  &1&numerical&\multicolumn{2}{|l|}{\cite{meteostat_weathers_2022}}&\checkmark& \checkmark  \\
Average windspeed & in km/h& city-wide&daily&daily \& hourly &1   $\mid$ 2&numerical&\multicolumn{2}{|l|}{\cite{meteostat_weathers_2022}}&\checkmark& \checkmark  \\
Average sea-level air pressure & in hPa & city-wide&daily&daily \& hourly &1   $\mid$ 2 &numerical&\multicolumn{2}{|l|}{\cite{meteostat_weathers_2022}} &\checkmark& \checkmark  \\
Sunshine duration &in minutes &city-wide &daily& &1&numerical&\cite{meteostat_weathers_2022}& &\checkmark&   \\
Peak wind gust & in km/h&city-wide&daily& &1&numerical&\cite{meteostat_weathers_2022}&& \checkmark&  \\
Relative humidity & \%&city-wide&& hourly &1&numerical&  &\cite{meteostat_weathers_2022}& & \checkmark  \\
Dew point & in C°&city-wide&& hourly &1&numerical&  &\cite{meteostat_weathers_2022}& & \checkmark  \\

  \hline
   \multicolumn{11}{|l|}{Graph connectivity indicators}\\
 \hline
 Degree& see appendix \ref{apx:connectivity_measures} &street-level&\multicolumn{2}{|l|}{constant}&1&numerical& computed based on \cite{senate_department_for_the_environment_mobility_consumer_and_climate_protection_berlin_radverkehrsnetz_2024}&computed based on \citep{boeing_modeling_2025} & \checkmark &\checkmark \\
 Betweenness& see appendix \ref{apx:connectivity_measures}  &street-level&\multicolumn{2}{|l|}{constant}&1&numerical&computed based on \cite{senate_department_for_the_environment_mobility_consumer_and_climate_protection_berlin_radverkehrsnetz_2024}& computed based on \citep{boeing_modeling_2025}& \checkmark &\checkmark  \\
 Closeness&  see appendix \ref{apx:connectivity_measures} &street-level&\multicolumn{2}{|l|}{constant}&1&numerical&computed based on \cite{senate_department_for_the_environment_mobility_consumer_and_climate_protection_berlin_radverkehrsnetz_2024}& computed based on \citep{boeing_modeling_2025}& \checkmark &\checkmark  \\
 Clustering coefficient& see appendix \ref{apx:connectivity_measures}  &street-level&\multicolumn{2}{|l|}{constant}&1&numerical&computed based on \cite{senate_department_for_the_environment_mobility_consumer_and_climate_protection_berlin_radverkehrsnetz_2024}& computed based on \citep{boeing_modeling_2025}& \checkmark &\checkmark  \\
 Whether sensor is in main cycling network&&street-level&\multicolumn{2}{|l|}{constant}&1&binary& computed based on \cite{senate_department_for_the_environment_mobility_consumer_and_climate_protection_berlin_radverkehrsnetz_2024}& & \checkmark &  \\
  \hline
  \multicolumn{11}{|l|}{Holiday indicators}\\
  \hline
 Public holiday& & city-wide&\multicolumn{2}{|l|}{daily} &1&binary&\cite{senate_department_for_education_youth_and_family_ferientermine_2023}& \citep{new_york_city_department_of_education_2011-12_2017} & \checkmark &\checkmark  \\
 School holiday & &city-wide &\multicolumn{2}{|l|}{daily} &1&binary&\cite{senate_department_for_education_youth_and_family_ferientermine_2023}&\citep{new_york_city_department_of_education_2011-12_2017} & \checkmark &\checkmark  \\

 \hline
 \multicolumn{11}{|l|}{Time indicators}\\
 \hline
 Hour& $\in$ \{0,$\cdots$, 23\} &city-wide&&hourly &1&categorical & &inherent& &\checkmark \\
Day of the week& Monday, Tuesday, ... &city-wide&\multicolumn{2}{|l|}{daily} &1&categorical & \multicolumn{2}{|l|}{inherent}& \checkmark &\checkmark \\
 Month& &city-wide&\multicolumn{2}{|l|}{monthly} &1&categorical &\multicolumn{2}{|l|}{inherent}& \checkmark &\checkmark \\
 Year & &city-wide &\multicolumn{2}{|l|}{yearly} &1&categorical & \multicolumn{2}{|l|}{inherent}& \checkmark &\checkmark \\
 Weekend& &city-wide &\multicolumn{2}{|l|}{daily} &1&binary & \multicolumn{2}{|l|}{inherent}& \checkmark &\checkmark \\

\hline
   \multicolumn{11}{|l|}{Motorized traffic indicators }\\
 \hline
 Average velocity of motorized vehicles, of cars, of lorries & in km/h&each within 6km radius around the sensor and within the whole city&daily&&6&numerical&\cite{berlin_open_data_verkehrsdetektion_2024}& & \checkmark & \\ 
Quantity of motorized vehicles, of cars, of lorries& &each within 6km radius around the sensor and within the whole city&daily&&6&numerical&\cite{berlin_open_data_verkehrsdetektion_2024}& & \checkmark & \\ 

    \hline
     \multicolumn{11}{|l|}{Socioeconomic indicators}\\
     \hline
Number of inhabitants& &planning area &yearly for 2019, 2020, constant for 2021, 2022, 2023&&1&numerical&\cite{berlin-brandenburg_office_of_statistics_kommunalatlas_2023}& &\checkmark& \\  
Proportion of inhabitants who have lived at the same address for 5 years or more & &planning area &yearly for 2019, 2020, constant for 2021, 2022, 2023&&1&numerical&\cite{berlin-brandenburg_office_of_statistics_kommunalatlas_2023}& &\checkmark& \\ 
Share of under 18-year-olds in the total population & &planning area &yearly for 2019, 2020, constant for 2021, 2022, 2023&&1&numerical&\cite{berlin-brandenburg_office_of_statistics_kommunalatlas_2023}& &\checkmark& \\
Proportion of 65+ in relation to total population & &planning area &yearly for 2019, 2020, constant for 2021, 2022, 2023&&1&numerical&\cite{berlin-brandenburg_office_of_statistics_kommunalatlas_2023}&&\checkmark & \\
Average age & &planning area &yearly for 2019, 2020, constant for 2021, 2022, 2023&&1&numerical&\cite{berlin-brandenburg_office_of_statistics_kommunalatlas_2023}& &\checkmark& \\
Proportion of people with a migration background in relation to the total population  & &planning area &yearly for 2019, 2020, constant for 2021, 2022, 2023&&1&numerical&\cite{berlin-brandenburg_office_of_statistics_kommunalatlas_2023}& &\checkmark& \\
Proportion of foreign nationals in the total population  (total, EU-foreigners, non-EU-foreigners)& &planning area &yearly for 2019, 2020, constant for 2021, 2022, 2023&&3&numerical&\cite{berlin-brandenburg_office_of_statistics_kommunalatlas_2023}& &\checkmark& \\
Gender distribution & &planning area &yearly for 2019, 2020, constant for 2021, 2022, 2023&&1&numerical&\cite{berlin-brandenburg_office_of_statistics_kommunalatlas_2023}&&\checkmark & \\
Fertility rate & &planning area &yearly for 2019, 2020, constant for 2021, 2022, 2023&&1&numerical&\cite{berlin-brandenburg_office_of_statistics_kommunalatlas_2023}&&\checkmark & \\
Share of unemployed & &planning area &yearly for 2019, 2020, constant for 2021, 2022, 2023&&1&numerical&\cite{berlin-brandenburg_office_of_statistics_kommunalatlas_2023}&&\checkmark & \\
Youth quotient & &planning area &yearly for 2019, 2020, constant for 2021, 2022, 2023&&1&numerical&\cite{berlin-brandenburg_office_of_statistics_kommunalatlas_2023}&&\checkmark & \\
Age quotient & &planning area &yearly for 2019, 2020, constant for 2021, 2022, 2023&&1&numerical&\cite{berlin-brandenburg_office_of_statistics_kommunalatlas_2023}&&\checkmark & \\
Greying index & &planning area &yearly for 2019, 2020, constant for 2021, 2022, 2023&&1&numerical&\cite{berlin-brandenburg_office_of_statistics_kommunalatlas_2023}&&\checkmark & \\
Average migration volume ( people moving to plus people moving away per 100 inhabitants)  & &planning area &yearly for 2019, 2020, constant for 2021, 2022, 2023&&1&numerical&\cite{berlin-brandenburg_office_of_statistics_kommunalatlas_2023}&&\checkmark & \\
Average net migration (people moving there minus people moving away per 100 inhabitants)  & &planning area &yearly for 2019, 2020, constant for 2021, 2022, 2023&&1&numerical&\cite{berlin-brandenburg_office_of_statistics_kommunalatlas_2023}&&\checkmark & \\
 \hline
 Additional indicators & &&&&&\\
 \hline
Whether the sensor measurement is missing & & street-level&\multicolumn{2}{|l|}{constant}&1&binary&\multicolumn{2}{|l|}{inherent}& \checkmark &\checkmark \\
Street ID &  & street-level&\multicolumn{2}{|l|}{constant}&1&categorical& \multicolumn{2}{|l|}{inherent} & \checkmark &\checkmark  \\
\hline
&&&&&&&&&$\sum$ 98 features&$\sum$ 79 features\\
\bottomrule
    \end{longtable}
\end{tiny}

\end{landscape}
\section{Feature Importance Analysis}\label{apx:feature_importance}

\begin{table}[h!]
\centering
\tiny
\caption{Feature Importance Groups for Taxi and Strava Datasets (values scaled by 10,000). Scores of one-hot encoded categorical features and features computed at multiple radii (e.g., hospital counts within 0.05–0.5 km) are aggregated using the absolute sum.}
\label{tab:feature_importance_combined}
\begin{subtable}[t]{0.95\textwidth}
\centering
\caption{Taxi Dataset}
\begin{tabular}{|l|c|c|c|p{6cm}|}
\hline
\textbf{Group Name} & \textbf{Mean} & \textbf{Sum} & \textbf{No° Features in Group} & \textbf{Features (Importance)} \\
\hline
Infra. (Built) & 1048.92 & 9440.26 & 9 & street type (3975.59), street surface (2310.95), number of street lanes (1738.58), max speed (580.54), longitude (341.46), distance city center (170.93), latitude (138.44), is street one way (93.78), parking (89.99) \\
\hline
Infra. (POI) & 698.52 & 7683.71 & 11 & count hospitals (1171.37), count hotels (1160.42), count shops (1017.01), count industry (834.12), count bars (752.56), count education (721.87), count restaurants (614.71), count subway lightrailway (485.64), count bus stops (443.02), count long-distance bus stops (259.23), count cafes (223.77) \\
\hline
Target Variable & 288.51 & 288.51 & 1 & target feature (288.51) \\
\hline
Connectivity & 241.36 & 965.45 & 4 & degree (318.97), closeness (240.02), betweenness (210.35), clustering (196.12) \\
\hline
Weather & 165.26 & 2148.44 & 13 & pressure (284.43), snowfall (271.74), hourly precipitation (246.82), temperature avg. (246.62), hourly wind speed (194.61), hourly relative humidity (193.12), temperature max (192.14), precipitation (140.94), temperature min (95.62), wind speed avg. (88.93), hourly dew point (88.49), hourly pressure (67.92), hourly temperature (37.05) \\
\hline
Features (Missing/Masked) & 120.00 & 240.00 & 2 & binary feature indicating missing (142.77), binary feature indicating masked (97.24) \\
\hline
Holidays & 24.89 & 49.79 & 2 & public holiday (32.41), school holiday (17.38) \\
\hline
Time & 19.19 & 76.77 & 4 & day of week (33.17), month (24.21), hour (13.93), weekend (5.46) \\
\hline
\end{tabular}
\end{subtable}

\vspace{0.5cm}

\begin{subtable}[t]{0.95\textwidth}
\centering
\caption{Strava Dataset}
\begin{tabular}{|l|c|c|c|p{6cm}|}
\hline
\textbf{Group Name} & \textbf{Mean} & \textbf{Sum} & \textbf{No° Features in Group} & \textbf{Features (Importance)} \\
\hline
Time & 1017.80 & 4071.21 & 4 & month (1945.71), year (1143.46), day of week (789.43), weekend (192.60) \\
\hline
Infra. (POI) & 481.10 & 2405.51 & 5 & count education (771.11), count hotels (623.53), count hospitals (395.17), count shops (369.08), count industry (246.62) \\
\hline
Infra. (Built) & 238.10 & 6428.72 & 27 & street type (1920.03), street surface (1685.42), bicycle lane type (537.04), cyclability touring (467.14), max speed (330.28), is within cycling route (236.52), horticulture percent (210.71), cyclability commute (186.05), public facilities percent (168.03), arable land percent (115.82), latitude (103.64), number of street lanes (82.91), commercial area percent (54.69), grassland percent (51.94), weekend house area percent (51.58), water bodies percent (34.13), cyclability (27.73), longitude (26.18), distance city center (23.51), city square percent (22.88), residential use percent (22.62), waste disposal percent (17.19), allotment gardens percent (16.48), park area percent (16.45), forest area percent (10.17), traffic area percent (9.54), cemetery percent (0.04) \\
\hline
Features (Missing/Masked) & 179.12 & 358.23 & 2 & binary feature indicating missing (219.11), binary feature indicating masked (139.12) \\
\hline
Socioeconomic & 165.45 & 2812.62 & 17 & net migration per 100 (374.00), migration volume per 100 (361.02), gender distribution (307.00), share foreign eu nationals (252.13), greying index (232.83), old age dependency ratio (204.42), unemployment rate age 15 to 65 (157.57), share under 18 (152.26), share with migration background (118.03), share foreign non eu nationals (114.50), total fertility rate (111.94), youth dependency ratio (95.57), share foreign nationals (90.54), share 65+ (79.70), residents 5+ years same address (76.69), average age (50.31), total population (34.09) \\
\hline
Target Variable & 86.47 & 86.47 & 1 & target feature (86.47) \\
\hline
Connectivity & 61.30 & 306.51 & 5 & main cycling network (280.18), clustering (22.01), betweenness (1.86), closeness (1.50), degree (0.95) \\
\hline
Weather & 40.20 & 361.79 & 9 & temperature max (151.93), wind speed avg. (69.50), pressure (46.22), temperature avg. (35.51), snowfall (26.84), wind speed gust (13.31), temperature min (12.09), sunshine duration (3.34), precipitation (3.04) \\
\hline
Holidays & 39.29 & 78.59 & 2 & public holiday (72.79), school holiday (5.79) \\
\hline
Motorized & 6.82 & 81.86 & 12 & avg. speed cars (21.68), avg. speed all vehicles (20.40), avg. speed trucks (17.62), vehicle count all vehicles (4.85), avg. vehicle speed 6km (4.78), vehicle count trucks 6km (4.55), vehicle count cars (4.50), avg. speed cars 6km (1.21), vehicle count cars 6km (1.01), avg. speed trucks 6km (0.74), vehicle count all vehicles 6km (0.30), vehicle count trucks (0.22) \\
\hline
\end{tabular}
\end{subtable}
\end{table}

To improve interpretability and support future data collection, we conduct a feature importance analysis using \acrlong{ig} (\acrshort{ig}).

\acrshort{ig} is not only compatible with graph neural networks but particularly well-suited to our setting, as its additive nature enables straightforward aggregation of attribution scores. Since GNNUI operates on temporal windows where the same features are repeated across multiple time steps per node, this property allows meaningful aggregation of attributions across both temporal (time window) and spatial (nodes) dimensions. 

We compute \acrshort{ig} using 50 interpolation steps and average the gradients along the path from baseline to input. Scores are then aggregated across time and space using the absolute sum. Finally, we merge one-hot encoded categorical features and features computed at multiple radii (e.g., hospital counts within 0.05–0.5 km).

Table  \ref{tab:feature_importance_combined} reports both individual and grouped feature importance scores (mean and sum) for the Strava and taxi data. We include group-level scores because many features are obtained together (e.g., weather variables), so identifying key feature groups helps guide future data collection. For both Strava and taxi data, infrastructure features dominate, including both built infrastructure and points of interest. Street type, street surface, and mode-specific infrastructure are particularly influential, with the number of lanes and maximum speed for taxis, and the bicycle lane type for Strava. However, the datasets show differences in their secondary drivers: for taxi data, the target variable and connectivity measures show medium importance, while for Strava, socioeconomic features and indicators for missing/masked data are relevant. Most notably, temporal features exhibit strikingly different importance patterns between the datasets. Time indicators are crucial for Strava (with month and year having the highest scores), but much less critical for taxi data. This difference likely reflects the distinct temporal characteristics of the datasets: taxi data consists of hourly observations across two months, while Strava data comprises daily observations across four years, capturing much broader temporal variation. Additionally, cycling behavior may inherently exhibit greater temporal variability than taxi usage, with seasonal patterns and recreational cycling trends creating more pronounced time-dependent effects.

It is important to note that these findings are closely tied to the estimation of Strava and taxi data specifically. While general tendencies are observable, the relative importance of features may differ when modeling overall cycling or motorized traffic volumes. This is especially relevant given the known biases in crowdsourced data. Strava users, for instance, tend to be younger, male, and potentially more influenced by amenities like restaurants than the average cyclist. For a broader feature importance analysis of general cycling patterns, refer to \citep{kaiser_counting_2025}.

\section{Connectivity Measures}\label{apx:connectivity_measures}

The degree specifies the number of edges connected to a node $v_i$: 

\[
\text{degree}(v_i) = \text{number of edges incident to } v_i.
\]

Betweenness centrality of a node $v_i$, indicates the extent to which it lies on the shortest path between pairs of other nodes $v_j$ and $v_p$. With $\sigma_{v_jv_p}$ the total number of shortest paths from $v_j$ to $v_p$ and $\sigma_{v_jv_p}(v_i)$ the number of those paths passing via $v_i$,

\[
\text{betweenness}(v_i) = \sum_{v_j \neq v_i \neq v_p} \frac{\sigma_{v_jv_p}(v_i)}{\sigma_{v_jv_p}}.
\]

Closeness of a node, indicates how quickly all other $n$ nodes can be reached in the graph, with $d(v_i,v_j)$ the shortest path distance between $v_i$ and $v_j$,

\[
\text{closeness}(v_i) = \frac{1}{\frac{1}{n-1} \sum_{v_i \neq v_j} d(v_i, v_j)}.
\]


The clustering coefficient measures the degree to which nodes in the neighborhood of node $v_i$ are connected, providing insight into the local density. With the number of triangles centered at $v_i$ referring to the closed loops of length 3 that include node $v_i$ and its neighbors:

\[
C(v_i) = \frac{2 \times \text{number of triangles centered at } v_i}{\text{degree}(v_i) \times (\text{degree}(v_i) - 1)}.
\]
\section{Combination of Adjacency Matrices \label{appx:different_adjacency_matrices}}

Since binary and distance-based adjacency matrices yield promising results, we also investigate whether combining these approaches further enhances performance. To facilitate this, we adapt the architecture of GNNUI.

\begin{figure}[!ht]
    \centering
    \includegraphics[width=0.5\textwidth]{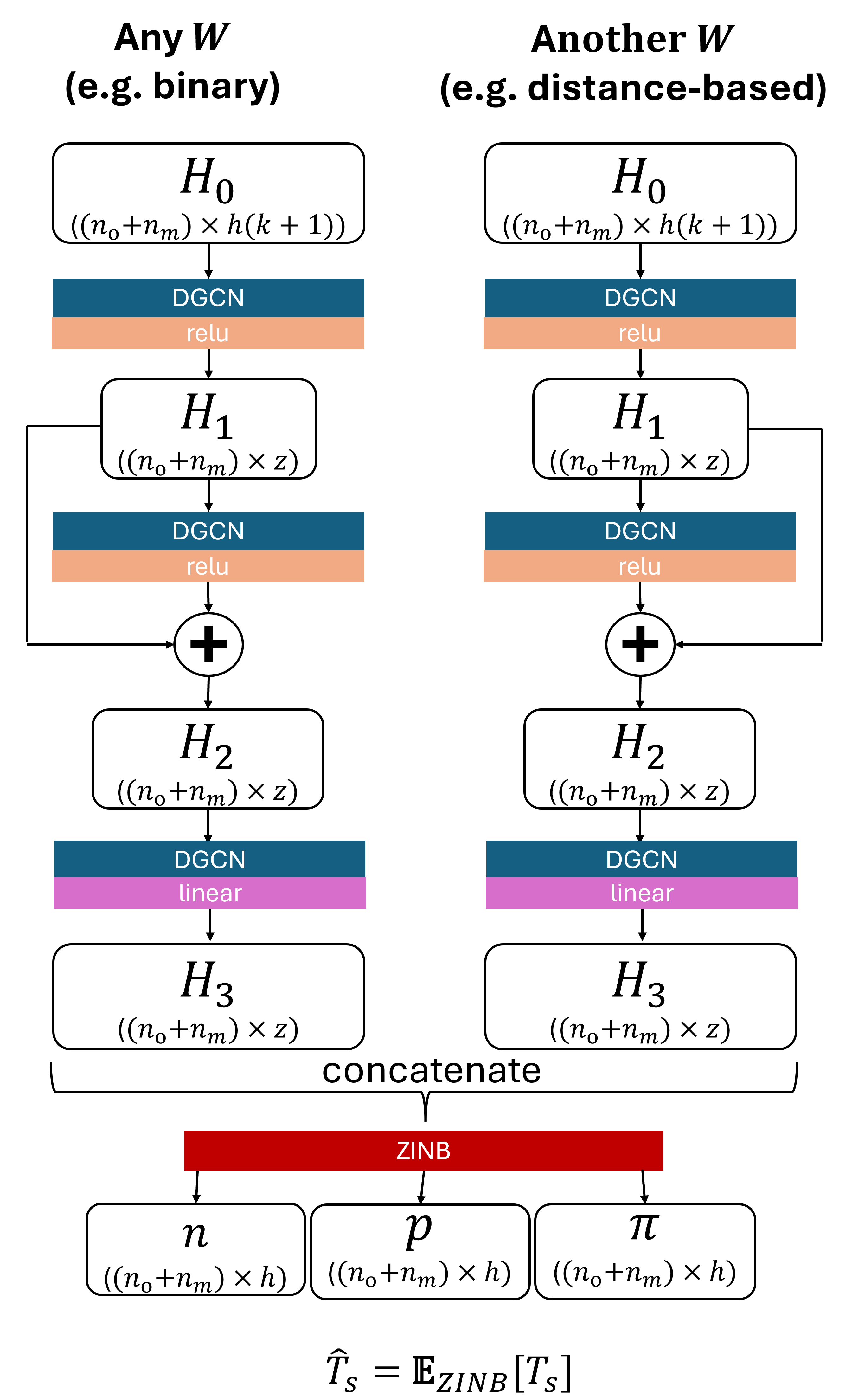}
    \caption{The pipeline architecture, when using a combination of adjacency matrices. }
    \label{fig:pipeline_twoA}
\end{figure}

We only explain the adapted architecture of GNNUi (with a usage of \acrshort{zinb} loss), as only this case is employed here. However, the procedure for the other losses is the same and can be transferred accordingly. For both adjacency matrices, we seperately proceed with the same procedure as if we were to use only one adjacency matrix, passing the input through the three DGCN layers, using a relu activation function after the first two and a linear one after the last one to obtain $H_3 \in \mathbb{R}^{n_o+n_m \times z}$. We then concatenate those along the second dimension to $H_3,c \in \mathbb{R}^{n_o+n_m \times 2z}$. This is then passed through a ZINB layer to obtain $n$, $p$, and  $\pi$. This is depicted in Figure \ref{fig:pipeline_twoA}. Because the subnetworks for each adjacency matrix are trained separately up to H3, this approach results in a substantial increase in computational requirements.

Table \ref{tab:compare_adjacency_matrices_twoA} presents the performance results of GNNUI, given the different combinations of adjacency matrices. Models that combine binary adjacency with another type perform well, likely benefiting from the strong performance of the binary structure. However, none of these combinations outperforms the standalone binary adjacency matrix, except for the KL for the taxi data (Binary is at 0.731 and Binary \& Similarity at 0.744). Due to the small numerical difference, we think this is negligible. Interestingly, the combination of similarity and distance-based adjacency matrices improves upon the individual specifications in certain cases. For example, in New York, the Similarity + Distance Time combination achieves an MAE of 30.39, improving upon similarity (40.50) and distance time (31.68). While this suggests that hybrid approaches may be beneficial, they still fall short of the binary adjacency matrix. This may be because combining different adjacency matrices (e.g., binary and similarity) can introduce overlapping or weakly relevant connections, causing the model to be distracted by noisy or less informative edges and thereby diluting the strong local signal that the binary adjacency effectively captures.\\
 
\begin{table}[!ht]
\centering
\scriptsize
    \caption{Comparison of GNNUI given the combination of adjacency matrices for Within Period-Generalization.} %
  \label{tab:compare_adjacency_matrices_twoA}%
    \begin{tabular}{ll|rrrrr|rrrr}
        \toprule
        \multicolumn{1}{c}{Adjacency matrix}&   &   \multicolumn{4}{c}{Strava Berlin}& & \multicolumn{4}{c}{Taxi New York} \\
        \cline{3-11} 
        &         & MAE &RMSE & KL & Zero & & MAE &RMSE & KL & Zero  \\
        \midrule
\midrule
Binary \& Similarity &   & 7.53 & 14.5 & 0.10 & 0.579 &   & 24.21 & 60.4 & 0.12 & 0.744 \\
Binary \& D. Birdfly &   & 7.61 & 14.4 & 0.09 & 0.547 &   & 29.49 & 64.7 & 0.10 & 0.672 \\
Binary \& D. Street &   & 7.97 & 14.9 & 0.09 & 0.505 &   & 29.04 & 63.6 & 0.08 & 0.694 \\
Binary \& D. Time &   & 7.71 & 15.1 & 0.01 & 0.577 & & 29.18 & 64.3 & 0.09 & 0.677 \\
\midrule
Similarity \& D. Birdfly &   & 9.18 & 16.9 & 0.14 & 0.428  &  & 30.60 & 67.7 & 0.20 & 0.565 \\
Similarity \& D. Street &   & 8.99 & 16.4 & 0.10 & 0.436 &   & 30.80 & 71.5 & 0.27 & 0.549 \\
Similarity \& D. Time &   & 9.29 & 16.9 & 0.12 & 0.449 &   & 30.39 & 73.1 & 0.30 & 0.545 \\
        \bottomrule
        \end{tabular}
\end{table}

\section{Varying Percentage of Street Segments: For Similarity and Distance Based Adjacency Matrices \label{appx:percentages}}

\begin{figure}[ht!]
    \centering
    \begin{subfigure}{0.49\linewidth}
        \includegraphics[width=\linewidth]{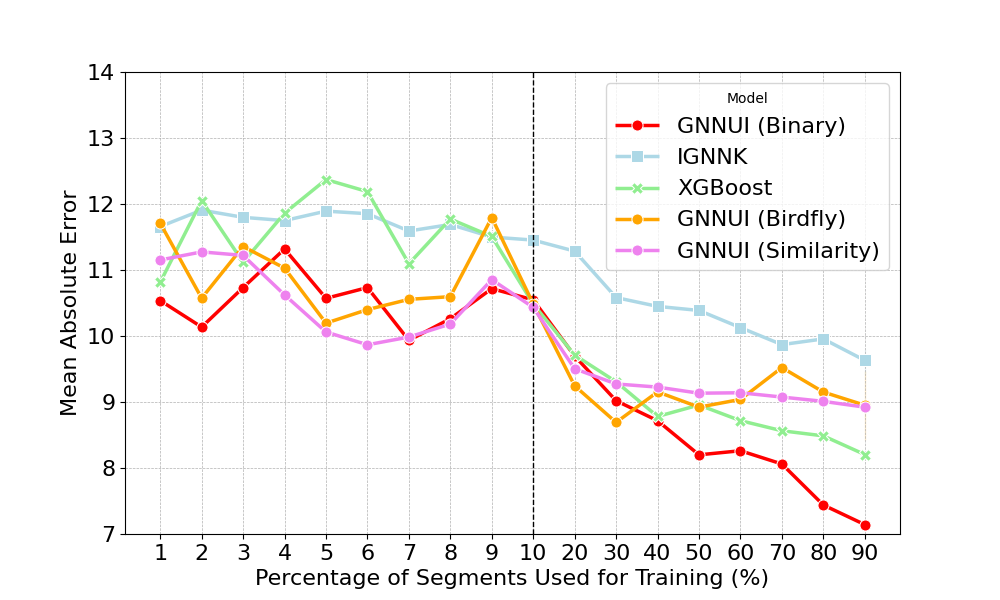}
        \caption{Strava}
        \label{fig:percentage_performance_Strava_appx}
    \end{subfigure}
    \hfill
    \begin{subfigure}{0.49\linewidth}
        \includegraphics[width=\linewidth]{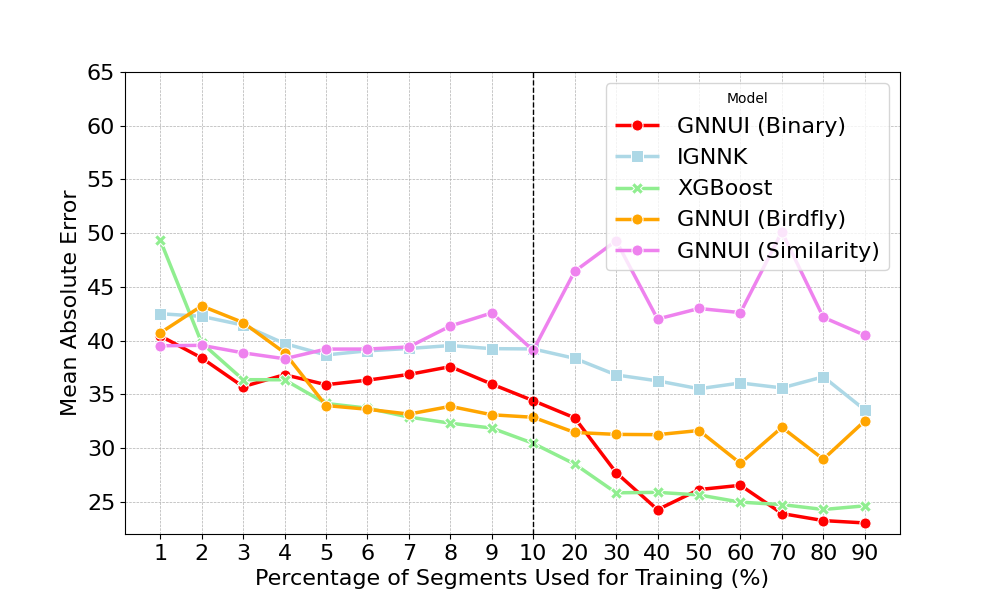}
        \caption{Taxi}
        \label{fig:percentage_performance_Taxi_appx}
    \end{subfigure}
    \caption{Performance of IGNNK, XGBoost and GNNUI (with binary, distance (birdfly) and similarity based adjacency matrix) under varying levels of data scarcity. Be aware that there is a structural break on the x-axis at 10\%.}
    \label{fig:percentage_performance_appx}
\end{figure}

\newpage
\bibliographystyle{elsarticle-harv}
\bibliography{references}

\end{document}